# Semantic Modelling of Organisational Knowledge as a Basis for Enterprise Data Governance 4.0 - Application to a Unified Clinical Data Model

**Miguel AP Oliveira[1], Stephane Manara[1], Bruno Molé, Thomas Muller, Aurélien Guillouche, Lysann Hesske, Bruce Jordan, Gilles Hubert, Chinmay Kulkarni, Pralipta Jagdev and Cedric R. Berger[2]**

## Abstract

Individuals and organizations cope with an always-growing amount of data, which is heterogeneous in its contents and formats. An adequate data management process yielding data quality and control over its lifecycle is a prerequisite to getting value out of this data and minimising inherent risks related to multiple usages. Common data governance frameworks rely on people, policies, and processes that fall short of the overwhelming complexity of data. Yet, harnessing this complexity is necessary to achieve high-quality standards. The latter will condition any downstream data usage outcome, including generative artificial intelligence trained on this data. In this paper, we report our concrete experience establishing a simple, cost-efficient framework that enables metadata-driven, agile and (semi-)automated data governance (i.e. Data Governance 4.0). We explain how we implement and use this framework to integrate 25 years of clinical study data at an enterprise scale in a fully productive environment. The framework encompasses both methodologies and technologies leveraging semantic web principles. We built a knowledge graph describing avatars of data assets in their business context, including governance principles. Multiple ontologies articulated by an enterprise upper ontology enable key governance actions such as FAIRification, lifecycle management, definition of roles and responsibilities, lineage across transformations and provenance from source systems. This metadata model is the keystone to data governance 4.0: a semi-automatised data management process that considers the business context in an agile manner to adapt governance constraints to each use case and dynamically tune it based on business changes.

[1] Co-first authorship; [2] Corresponding author
When this work was conducted, all authors were employed by Novartis Pharma AG with a direct or contractor contract.

**Contacts:** Miguel AP Oliveira https://ch.linkedin.com/in/oliveiramiguel; Stephane Manara https://www.linkedin.com/in/stephane-manara-fr68/; Bruno Molé https://www.linkedin.com/in/-bruno-mole-/; Thomas Muller https://www.linkedin.com/in/thomas1mueller/ Aurélien Guillouche https://www.linkedin.com/in/aur%C3%A9lien-guillouche-58234440/; Lysann Hesske https://www.linkedin.com/in/lysannhesske-data-management/; Bruce Jordan https://www.linkedin.com/in/bruce-c-jordan/; Gilles Hubert https://www.linkedin.com/in/gilles-hubert-7aa9a915/; Chinmay Kulkarni https://www.linkedin.com/in/chinmay-kulkarni-45a87a1/; Pralipta Jagdev https://www.linkedin.com/in/pralipt-jagdev-498a46a7; Cedric R Berger www.linkedin.com/in/cedricberg

# 1. Introduction

Getting value out of an ever-growing and fast-changing amount of data is not easy. While data is collected for a specific purpose (one primary use), it might be used in data science and contribute to potentially unlimited analyses (many secondary uses). In this context, it is essential to document data and its context throughout its lifecycle; this information is referred to as metadata and should be inseparable from the data asset it describes. Having metadata contextualising data assets is a prerequisite for applying fit-for-purpose data governance. Indeed, the governance shall be adjusted to each data asset as per its context (the 5 Ws). With a richly described context, we can apply adequate governance that might bring more constraints than benefits.

Data Governance (DG) is a vast theoretical topic that can be implemented in various manners. Based on recent literature reviews (1)(2)(3), we consider DG in its most generic definition: "A data management framework to get the maximum value out of data while minimising associated risks". Yet, implementing DG is not trivial, especially in big organizations (4), as it is a collective action problem (5)**.**

Traditional ways to implement DG rely on people following a set of official documents to apply a central, top-down, bureaucratic, restrictive and regulation-driven process. Authoritative governance bodies define and describe principles in documents that usually form a hierarchy of guiding documents (policies, standards, procedures, guidelines, etc.) explaining how data should be handled. Data expert roles apply these principles and enforce them at different organisational levels (6). Principles delineated in documents can also be used to configure DG tools. Multiple commercial solutions are proposing DG capabilities (7). Each solution has PROs and CONs and requires a significant monetary investment upfront for installation and integration in the existing IT ecosystem. Although these DG systems continuously improve, adopting them means adopting their limitations (8). Despite many open standards, DG solutions vendors lock in customers by using proprietary data formats or transformation processes (9).

Efforts to standardise DG are ongoing (10)(11), and the advantages it may bring to big organizations heavily depend on how it is implemented (12)(13), especially in the healthcare domain (14)(15). A one-size-fits-all DG solution is likely not to exist (16). Legacy DG, driven by a dedicated organization (people) promoting data practices explained in documents, is slowly reactive by nature. There is a need for modern DG where macro governance (at industry and enterprise levels) articulates dynamically with micro governance (at team and project levels) and where inherent governance constraints are adapted to business needs (17). In this paper, we introduce the concept of Data Governance 4.0: an agile (adapts dynamically), fit-for-purpose (adequate to business needs) and (semi)automated governance of data assets. DG4.0 application is driven by metadata describing the business context around data assets. Knowing data in its context enables us to balance the constraints of DG, adapt them when the context changes, and evaluate the overall benefits of DG (18)(19).

In this paper, we first describe the challenges associated with legacy DG. We provide background information and briefly review known solutions. In the second part, we introduce DG4.0, a new approach leveraging open standards and solutions to render DG agile and fit-for-purpose to business needs and automate its application. We explain the methodology we use to capture the knowledge about data assets and their business context and the technological stack supporting DG4.0. In the last part of the paper, we provide concrete examples of applying DG4.0 to manage the unification of 20 years of clinical study data in one target model. Eventually, we discuss the advantages of DG4.0 as compared to current solutions and how it can foster sustainable digital transformation for any organization.

# 2. Problem Statement

In 1968, Conway postulated that "organizations, who design systems, are constrained to produce designs which are copies of the communication structures of these organizations" (20). Applying it to data implies that "any (complex) organization that designs a system will produce a (complex) design" (21). The bigger the organization, the higher the data complexity; thus, DG could bring more benefits (22). Nevertheless, implementing DG in big and complex organizations remains a challenge from a conceptual and operational point of view. We list below three paradoxes inherent to DG and challenges we faced when implementing DG in our organization:

Intrinsic DG paradoxes:

1. Governance comes with constraints. Who is willing to pay for constraints? The cost of implementing governance is significant; if a budget is not explicitly dedicated to DG implementation, it will probably not be used for this aim.
2. DG is an enterprise endeavour; it is suboptimal to deploy it in pockets of an organization only. DG's horizontal (platform) nature is incompatible with the big organisation's legacy vertical (hierarchical) structure.
3. While DG benefits are global, constraints are local. Business experts creating data and contributing to the DG effort differ from those directly benefiting from DG outcomes, primarily for data integration globally.

DG challenges related to digital awareness:

4. If everybody agrees upon the theoretical benefits of DG, it is hard to quantify its benefits (such as return-on-investments). Implementing a traditional DG framework (processes and policy-driven) requires a significant upfront effort/investment; it is hard to demonstrate a measurable short or mid-term benefit. There is a need for reliable DG metrics.
5. Over-optimistic marketing promises from companies selling DG solutions create high expectations that are difficult to satisfy in big organizations. These circumstances erode the trust needed to invest in DG over the long run. There is a need to raise the digital awareness of decision-makers and better manage stakeholder expectations.
6. DG initiatives are digital projects and require a specific management methodology[1]. Managers should be educated about it and apply it systematically.
7. DG is a dry topic dealing with data curation activities behind the digital scene. DG outcomes cannot be communicated with fancy visual graphics; hence, there is little attractiveness in an outcome-based rewarding system. There is a need to raise digital awareness that emphasizes the added value of DG.

DG challenges related to operations:

8. The data landscape is complex; it is hard to apprehend as a whole and keep up-to-date with its changes. There is a need to synthesise knowledge about data assets and spread it across the organization; this is a must-have for efficient DG.
9. Big enterprises have IT and business (non-IT) organizations featuring different mindsets and incentivization logic. DG, being at the boundary of IT and business, generates significant friction. Data being a shared (enterprise) asset, there is a need to change the corporate mindset (culture), hence the incentivization and rewarding process.
10. DG's Technological infrastructures and operational costs are increasing proportionally to the amount of data and metadata to govern. For reasons pertaining to business sustainability, it is essential to design a sober DG framework, resilient to changes (people, skills, technology, standard), that does not contribute to increasing the technical debt of an organization.

Considering these challenges, we sought better ways to implement DG in our organization and foster its adoption.

---

[1] https://www.pmi.org.in/DSandAIPlaybook/

# 3. Background information

## 3.1. Traditional Data Governance

People, processes and policies: Usual DG is driven by people, processes and policies. Once DG actors have been defined (authoritative bodies, roles and responsibilities), they work on implementing and monitoring processes consisting of identifying, characterising and managing data assets. This activity is supported by various technological solutions featuring diverse capabilities that are more or less compatible as provided by different vendors (23)(24). DG activities usually result in maintaining centralised artefacts such as a business glossary, a technical data dictionary and a data catalogue.

Single-shot and reactive: Traditional DG is reactive to risks that have become issues. Our experience tells us that, historically, applying DG policies is primarily a risk management practice (such as preventing the risk of not being compliant with regulations) triggered by an audit finding or a risk assessment exercise. As a result, legacy DG activities are mainly single-shot exercises driven by one specific need and targeting one specific data asset. These non-reusable approaches are currently being challenged by laws addressing data confidentiality and privacy globally, such as GDPR or auditing requirements from external bodies that are steadily increasing. For example, sharing source data and analysis data is getting from frequent. In this case, tracing back lineage in a consistent fashion is essential.

System-centric: Traditional data management activities emerged to support the design of single systems (and their interactions with partner systems) that typically handle data in well-known (and quite slowly changing) formats. Before the rise of big data, interest and focus on data as reusable assets were less prominent. Moreover, the growing interest in DG to support data reuse (embodied by the development of FAIR principles) has shifted the data expertise towards business experts: only subject-matter experts know when the data makes sense. This case is even more true when data gets copied from the primary source systems into big data management platforms capable of dealing with larger volumes, velocity and variability. An extra effort in data discovery and metadata enrichment is mandatory at that point.

## 3.2. Semantic Web

Semantic Web standards, set by the World Wide Web Consortium (W3C), define a set of formalisms to capture knowledge in a human- and machine-readable format (21)(25). These standards enable encoding semantics (meaning) within data assets, making them helpful in integrating heterogeneous sources of knowledge and annotating them (including extra information on a data source). Rich annotation is especially powerful in highlighting similarities or differences across data assets to disambiguate or unify them (27).

Semantic web standards leverage Uniform Resource Identifiers (URI), which represent entities uniquely and unambiguously and create statements about them using the RDF[2] standard. RDFS[3], a modelling language for RDF, provides a mechanism for describing a group of related resources using a common vocabulary. The OWL[4] language enables rich knowledge description using complex axioms as part of ontologies (28)(29). SPARQL[5] is the query language to retrieve information from semantic web standards stored in RDF-triples databases (triple store).

---

[2] Resource Description Framework

[3] Resource Description Framework Schema

[4] Ontology Web Language

[5] Simple Protocol and RDF Query Language

SKOS[6] and SHACL[7] are standards recommended by W3C to standardise semantic-related capabilities through shared structured vocabularies. SKOS is a simple controlled vocabulary for representing knowledge organization systems (such as thesauri, classification schemes, taxonomies and subject-heading systems) (30)(31) while SHACL is a language for validating RDF data against a set of conditions that is based on a structured vocabulary (32).

## 3.3. FAIR Principles

FAIR principles promote data Findability, Accessibility, Interoperability, and Reusability (33). The recipe to implement these principles is published openly (34). Its successful implementation within an organization allows rigorous management and stewardship of all "FAIRified" resources (35). Furthermore, it significantly boosts data management as part of enterprise digital transformation (36), triggering resource savings (33) while ensuring more consistent and reproducible data science (37).

In scientific research, the FAIR guidelines have generated a concrete added value and a significant return on investment (33)(38)(39). Despite challenges associated with their implementation (40), advantages steered several biopharmaceutical companies to publicly declare their interest in adopting these guidelines and methodologies (36). FAIR principles ease data management by providing machine-actionable metadata (33), which is a prerequisite to applying artificial intelligence (41) and helping big organizations increase the overall data quality, as well as scaling and speeding up the generation of new insights (36).

In this paper, we report on how semantic web standards are helpful to model and integrate enterprise data assets in a FAIR-compliant manner. Because semantic web standards propose URI as unique identifiers and enable rich metadata annotation of data assets, we choose RDF as a framework of choice to apply FAIR principles. We use this combination to describe data assets and their business context within the enterprise. Contextual knowledge has been vital for efficient governance (42). Several papers report the implementation of RDF to capture and model enterprise assets and to address some of the FAIR requirements (43–49).

## 3.4. Enterprise Architecture

Designing an enterprise architecture (EA) framework allows one to rigorously describe the structure of an organization, its components (business entities), the properties of those components and their relationships (50). TOGAF (51) and the Zackman framework (52)(53) are the most popular resources for using standard building blocks to describe EA.

While these standards usually serve to document EA, we propose using them for DG purposes and, more specifically, to describe the enterprise/business context surrounding data assets subject to governance. Data asset contextualisation is essential as "a piece of information is defined only by what it's related to, and how it's related" (54). Capturing the complexity of the enterprise architecture in which data assets are embedded is vital to applying fit-for-purpose governance.

---

[6] Simple Knowledge Organization System

[7] Shapes Constraint Language

# 4. Knowledge Modelling and the Future of Data Governance

In this article, we report our attempt to solve problems associated with legacy DG as listed in the Problem Statement section. We choose semantic web standards, i.e., the RDF-based semantic stack (55), as a tool of choice to improve governance of our data assets and FAIRify them as part of the DG process. The primary artefact resulting from our work is a human- and machine-readable knowledge graph (KG) containing the necessary information to solve DG paradoxes and challenges. Herein, we define version 4.0 of Data Governance (DG4.0) as a novel way to perform enterprise DG. Therefore, this approach aligns with the fourth industrial revolution (56), enabling a (meta)data-driven, automatised, adjusted and agile governance of data assets.

## 4.1. Enterprise Knowledge Graphs

Using semantic web standards to capture data knowledge results in a KG (57) that describes data assets and their business context. The KG is queried, and knowledge is retrieved to fulfil DG use cases. The results of queries are displayed in different formats (narrative, tabular, graphical) as part of enterprise resources. We describe these various artefacts in this paper's Enterprise Resources section.

Domain-specific knowledge instantiated using semantic web principles is called ontology (58). The latter are the building blocks of linked data (59). Large organizations (60) can leverage ontologies to improve the quality of their data (61). This paper explains how we developed and used an enterprise ontology as metadata to describe data assets, apply FAIR principles (62) and depict the business context using architecture standards (63). The section Data Governance 4.0 Applied to Unified Clinical Data Model focuses on a specific example of FAIRifying and unifying 25 years of legacy clinical data. We explain how DG4.0 enables an agile, semi-automated governance of secondary uses of clinical data.

As detailed in Section 3.3, FAIR principles promote using URIs and rich metadata to characterise data sets and their content. FAIR principles enable humans and machines to identify and understand what a dataset is about, how to find it, access its content, use it across multiple systems, and what the conditions are to reuse it. However, applying these principles uniquely to the datasets does not enable humans/machines to understand the different contexts in which these datasets are used, where they are coming from and where they are going. To do so, we must complete FAIR principles with a detailed description of business facets. A dataset "is defined [...] by what it's related to, and how it's related" (64). These "what" and "how" aspects refer to many enterprise assets related to data assets such as IT systems/applications, infrastructure technologies, business processes, business domains and enterprise divisions/ departments, information security aspects, business projects, funding, etc. in other words, the entire business architecture (63).

Our enterprise ontology covers both EA and DG domains; EA entities are derived from TOGAF (51) and Zackman framework (52) (53), while DG entities come from the DAMA DMBOK (65). The DMBOK is a reference for DG professionals and defines eleven Data Management Knowledge areas (66). Our ontology covers these knowledge areas and links EA and DG entities with relationships we have determined based on our needs.

## 4.2. Future of Data Governance

We can only govern data that we know. But what is it to really know data? Elementary knowledge about data can be estimated along four dimensions: meaning, history, modelling and mastery (67); additional dimensions can be added over time. The maturity grows over time along each dimension as one progresses toward DG4.0. As explained in the Problem Statement section, governance has constraints and shall not hinder business value delivery.

Because business is highly versatile, governance must adapt to enable reasonable freedom to operate with data while mitigating identified risks. To this aim, we need to know the data we are supposed to govern, but what does it really mean "to know data" (68)? What does this knowledge encompass? How do we use this knowledge to define, apply and adjust DG? We address these questions by proposing a metadata-driven, automated, and agile DG4.0 framework aligned with the definition of the fourth industrial revolution (56).

Big organizations are complex and fast-changing; obtaining and maintaining an exhaustive understanding of their data ecosystem is challenging. RDF-based linked data has already been proposed as a promising solution to improve DG (60,69) and mitigate associated risks (70) in such complex environments. For effective enterprise DG, the KG shall ontologies (71) about:

- data assets (datasets or concepts represented in these datasets)
- business context (around these data assets)
- data governance

Beyond a semantic glossary (72), such a KG can be compared to an enterprise repository serving as a metadata library (73) to describe data assets in their business architecture and prescribe governance requirements to humans or machines. Few articles use an ontologies-based data management approach (74) to support DG metadata FAIRification and quality (67). We claim that our framework also reduces the inherent paradoxes and challenges associated with DG for the following reasons:

1. **Human and machine compatibility:** URIs are human- and machine-readable. Therefore, they can be used as requirements (for humans) or configuration metadata (for machines) to semi-automatically apply governance principles.
2. **It auto-describes itself:** Each URI can resolve in a web page and become a URL describing itself, improving transparent knowledge exposure and leading to a more coherent cross-organisational understanding of data assets.
3. **Flexible, cost-efficient, enterprise-specific solution:** This framework is lean and modular, using open standards, a simple process and commonly used IT tools. It is ideal for proofs-of-concept/value at a small scale before deployment at the enterprise level.
4. **It shows the value it adds:** We transparently expose how business needs are bound to the solution the KG provides for each use case. We demonstrate the vital link between the added values (and constraints) DG brings to specific use cases. It also enables agile DG and adjustments if it is too tight.

## 4.3. Toward Metadata-driven, Agile and Automated Data Governance

Our strategy toward DG4.0 is first to implement a descriptive framework that creates, maintains, enriches and improves the representation of our enterprise data assets. As a second step, we create and continuously improve a DG ontology that links governance items to data entities they apply to. This paper focuses on collecting and integrating knowledge about data assets and their ecosystem, a prerequisite to DG4.0. Our KG depicts both "as-is" (current) and "to-be" (required) governance and serves as a descriptive or prescriptive resource to apply governance. The concrete application and monitoring of DG at the enterprise scale are not in the scope of this paper.

The following sections describe our methodology and the technical aspects of its implementation. Furthermore, the last section describes how we use our framework to unify twenty years of legacy clinical study data and how we apply governance.

# 5. Knowledge Management Methodology

Previous sections explain the need for governance metadata (KG) and a query language to retrieve knowledge from it. The metadata model aims at describing data assets and their business context. This "knowledge about data" is a prerequisite to inform humans and machines to perform an agile, fit-for-purpose and automated data governance, i.e. DG4.0. This section describes the methodology and processes we use to extract, integrate and expose data knowledge, including:

- knowledge elicitation from three different sources (humans, documents and machines)
- the knowledge integration from these sources
- the different enterprise resources resulting from it

The idea of having a central enterprise knowledge repository is not new and has been promoted by the Enterprise Knowledge Graph Foundation (75). This paper describes our hands-on experience building a KG on a use-case basis, project by project. We applied the same methodology for each project, which was composed of three main steps: knowledge elicitation, integration, and exposure. This section details how we collect knowledge and use semantic web principles (25) to integrate this knowledge and publish several enterprise resources supporting the management and governance of data assets.

For knowledge elicitation and transcription in our KG, we followed APQC guidelines (76–78). The method consists of predefined steps to capture tacit knowledge from experts and communities of practice (79). The gathered information was loaded in the KG using Excel templates featuring entities defined in our upper ontology.

The knowledge management methodology used in this work is composed of 3 main steps, as depicted within Figure 1, and detailed below:

1) Knowledge is retrieved from three types of sources: humans, documents and machines or data it contains.
2) Knowledge is then captured and integrated using business analysis with humans (steps 1,2, and 3), text entity extraction for documents (steps 4, 5 and 6) or data discovery/profiling for machines (steps 7, 8 and 9). The outcome is stored in RDF in the KG and linked to business needs expressed as Competency Questions. This information is linked to the existing content of the KG by reusing entities from the enterprise's upper ontology. The goal of this step #2 is to generate a semantic logical model with as much standard structure and content as possible.
3) We apply the governance principle to the logical model created in step #2. Important aspects of governance are the provenance and lineage of knowledge and the disambiguation and unification of logic with other digital entities or assets.

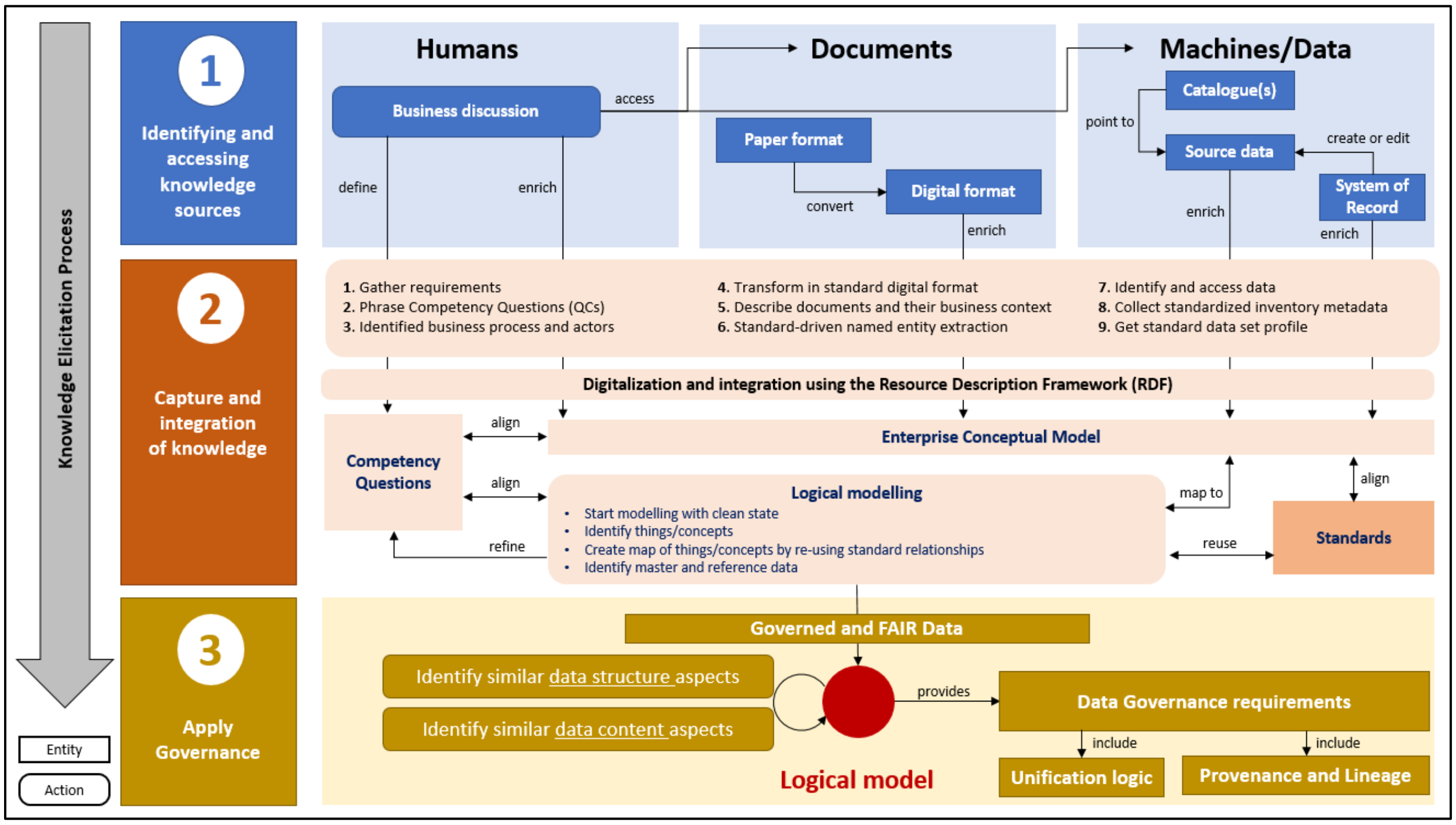

*Figure 1: Knowledge management methods. from knowledge elicitation, capture and integration to data governance.*

## 5.1. Knowledge Elicitation

Multiple knowledge elicitation techniques have been described in the literature (80), and we followed a structured approach to systematic elicitation (81). Our focus was to collect business needs, unanswered questions, user requirements or competency questions (CQ) (82) related to a specific project or use case. To interact with experts, we applied the APQC's "Knowledge transfer through peer-based approaches" (83), such as conversation/dialogue with experts, narratives and storytelling, Q&A Exchanges, webinars… During those sessions, we used upper ontology classes, properties, and individuals to capture knowledge as triple by translating experts' stories in triple structure (subject-predicate-object).

Figure 1 illustrates how we capture and concentrate information (coming in various shapes and from diverse sources) in the logical model (84). The latter represents information at adequate granularity to spread data democratisation (85).

### 5.1.1. Knowledge from experts

Most knowledge resides in employees' heads, commonly called "know-how". Such knowledge is not documented anywhere; hence, a knowledge elicitation process often starts with discussions with subject matter experts (SMEs). These discussions unfold as follows:

1. **Gather requirements:** let SMEs freely explain their problem with data
2. **Phrase competency questions (CQ):**
    a) Guide SMEs to phrase concise CQs or problem statements
    b) Identify main concepts and define them
    c) Reuse existing concepts if possible
3. **Identify business process and actors**

Collaboration with business experts is at the heart of knowledge elicitation. Continuous conversion of SMEs' inputs into RDF triples is the knowledge thread that links the CQs to solutions achieved via DG4.0. For instance, this thread is fundamental to supporting the generation of clean and linked data products (86) and assigning responsibilities as part of a data mesh (87). This step is also crucial for SMEs to identify all relevant resources to mine (e.g. documents).

It might sound trivial, but phrasing CQs at the right granularity level is an art and a crucial step in problem-solving. How one defines the problem determines whether you solve it (88). Markman explains in his article that retrieving pertinent information from graphs works the same way as for human brains: "When doing creative problem solving, the statement of the problem is the cue to memory. That is what reaches into memory and draws out related information". Relating/linking information smartly is our goal; therefore, it is essential to align CQs to conceptual and logical entities represented in the KG (Figure 1, part 2 "Capture and integration of knowledge"). Good CQs can be quickly reverse-engineered into SPARQL, which ultimately reflects the information structure within the logical model and dictates the ability to answer the query optimally. Suppose a query outcome delivers added value (e.g. new insights, resource savings, process improvement, data unification, data lineage, etc.) to business SMEs (Figure 1, part 3 "Apply governance"). In that case, we capture this feedback and link it to initial the requirements/CQs. We then refine existing CQs or create new ones, and then we iterate on this process and incrementally show the added value using agile/scrum ways of working (89)(90)(91).

### 5.1.2. Knowledge from documents

Knowledge extraction from documents or unstructured data is not the main topic of this paper. If needed, we perform the following sequential steps:

1. Transform documents into a standard digital format (e.g., HTML) using services from external vendors
2. Describe business context around documents (from discussions with SMEs)
3. Perform a guided named entity recognition and extraction using standard terminologies from our KG or from an external service provider

The section Content extraction from documents details how we process clinical study protocols to extract entities relevant to the unified clinical data model.

### 5.1.3. Knowledge from machines/data

Big organizations run many IT systems on servers or personal laptops. In our discussions with SMEs, we identify datasets, systems, and data transformations and articulate these concepts in precise QCs:

1. Identify and access data (sets), systems-of-records and data flows; ideally, get a sample of data
2. Collect standardized inventory metadata
3. Get a standardized profile for each dataset

## 5.2. Knowledge Integration

The integration of RDF triples acquired during the knowledge elicitation process on a project/use case basis integrate at a logical modelling level (84) in a logical schema (red dot in Figure 1) in the KG. The components of the KG logical schema result from the knowledge integration activities listed below and in Table 1).

**Business analysis:** As listed in the APQC's "Knowledge transfer through peer-based approaches", a normalised business analysis is performed for every new project. This analysis starts with conversation/dialogue with experts, narratives and storytelling, and Q&A Exchanges to collect information data assets and their business context (Table 1, case #1). This step leverages our enterprise upper ontology (UO, detailed further in the section Enterprise upper ontology), featuring enterprise architecture (EA) building blocks. Microsoft Excel templates (Excel Enrichers for business discussions) capture business knowledge systematically and consistently. These templates use entities defined by TOGAF (51) and the Zachman framework (52)(53). Business analysts and/or data experts populate these templates upon business analysis. Data-friendly business experts might populate Excel templates directly.

**Re-use of existing entities:** Upon knowledge elicitation, experts often mention identical or similar concepts (same problem, same important entity, same process bottleneck, etc.). To avoid re-creating concepts, we compare existing knowledge (class, property, individual) acquired for other projects to newly collected concepts. We compare new concepts to existing ones, and existing concepts are re-used. Doing so, we join different projects with concepts they share. Similar to OpenRefine (92)(93), we link or replace (if it applies) new concepts with existing knowledge. We thereby consolidate and increase the quality of enterprise knowledge (more precisely, consistency).

**Terminology management:** Upon business analysis, we collect business terms (business concepts) defined by business experts. We collect different definitions for the same given term to represent the heterogeneity of understandings across business cases. Each definition is tagged with its source: a human, a document or a machine/data set (Figure 1). We further annotate each term whenever possible by adding external definitions from open standards (e.g., Linked Open Data (94)). As part of our terminology management activities (Table 1, case #2), this process results in the iterative building of a business glossary: a flat list of URIs, labels (terms), synonyms and definitions. This ensures all terms used in our KG are richly and unambiguously defined.

**Information modelling:** Once captured and defined, terms from the business glossary can become modelling entities such as classes, objects, attributes, properties or individuals in an information model. We create information models (95) for knowledge collected from humans, documents or data/machines sources. We group under the term "information model" either data models (diagrams representing conceptual, logical, and physical for at-rest or in-motion) or any other EA models such as business process diagrams, system architecture diagrams, organization units diagrams, and interface diagrams. (Table 1, case #3). As detailed by Panos Alexopoulos (96), we use semantic data/information modelling (RDF-triples as modelling bricks) instead of traditional UML [https://en.wikipedia.org/wiki/Unified_Modeling_Language] entity-relationships modelling (97).

**Data FAIRification:** Terms from the business glossary can also annotate data assets (Table 1, case #4) for FAIRification. Using terms/concepts coming from open standards (DCAT (98), FOAF (99), Dublin Core (100)(101), PROV-O (102) ontologies), we tag datasets to fulfil each FAIR aspect for a given data asset.

1. Findability: A data asset is represented as a global, unique, persistent and resolvable identifier (URI in the KG) that resolves as a URL in an intranet page describing the data asset.
2. Accessibility: We link the data asset URI to terms/concepts describing what communication protocol to use and/or the procedure to access the data asset.
3. Interoperability: We link the data asset URI to terms/concepts describing the use of any standard knowledge representation, standard vocabulary and/or links to open reference data.
4. Reusability: We link the data asset URI to terms/concepts describing license terms, provenance or domain-specific standards.

**Master data management:** Terms from the business glossary can become master data (MD) or an MD attribute/property. For instance, the MD "Gene" has the attribute "Gene name," and both "Gene" and "Gene name" are listed and defined in the business glossary. From their RDF-triples version in the KG, MD and their attributes are exposed in standard format (XML/XSD, JSON-LD, TTL…) for global reuse across the organization (Table 1, case #5). The most important MD attributes are dedicated to capturing the multiple identifiers (usually non-human readable) given to the same MD object by different systems/applications. For thorough MD management, we apply SHACL constraints when capturing and reconciling these attributes.

**Reference data management:** Terms from the business glossary can either be the name of a reference list (flat list or taxonomy) or an item inside a list or a taxonomy (Table 1, case #6). Using SKOS relationships ("exactMatch"[8], "closeMatch"[9] or "narrowMatch"[10]), we perform inter- and intra-list mapping (30).

- Inter-list mapping: we map list-to-lists in a one-to-many manner
- Intra-list mapping: we map terms-to-terms in a many-to-many manner (across lists)

This process enables us to link data from different business domains, highlight the differences (at the list or term level) and reconcile entities if needed. This mapping activity is essential for scaling enterprise-wide mapping activities based on fuzzy logic, such as OpenRefine variable mappings (103).

**External data standard management:** Terms from the business glossary can either be specific to our organization or shared across the healthcare industry. Such terms are often available publicly or via license fees as standard terminologies. Health authorities request that pharmaceutical companies consistently use data standards to report domain-specific data. Because standards are numerous, overlap, and change at a different pace, we need a methodology to buffer this complexity in our organization (Table 1, case #7). Typically, we ETL or ELT external standards into a semantic format using our governed enterprise URI syntax. We add a timestamp for versioning. We often import only a subset of an external standard according to enterprise business needs and tag it with FAIR annotation such as licensing, provenance, etc… (See above Data FAIRification). Similar to reference lists (See above Reference data management), we perform inter- and intra-standard mapping. We process semantic web standards (public ontologies) in the same way.

**Data governance (DG):** Terms from the business glossary can become DG entities: a principle, a rule, a constraint, a check, a property, a class, an authoritative body or actor, an official document, etc. These DG terms are grouped in our KG as a "governance ontology" (Table 1, case #8). High-level governance principles link to official documents (regulatory policies or laws) from which rules or checks are derived. These rules/checks are linked to DG actors responsible for applying them and monitoring their application (DG process). Eventually, DG policies, rules/checks and activities/tasks are linked to the data entities they apply to (for instance, a data set, a data table, or an attribute as represented in one or many information models). We use the KG as a resource to describe and prescribe DG requirements. These requirements are human- and machine-readable and can, therefore, be applied by humans and/or machines, enabling DG automation.

**Metadata governance** has mostly to do with the management of our KG structure and content. As illustrated in Figure 2, we govern strictly URI syntax. It comprises predefined tags separated by "slash", as suggested in W3C recommendations (104). Each URI element (between two slashes) comes from a reference list from a single source of truth (105) managed by clearly identified actors. The URI syntax implemented in our KG is depicted in Figure 2 and it can be detailed as follows:

---

[8] https://www.w3.org/2009/08/skos-reference/skos.html#exactMatch

[9] https://www.w3.org/2009/08/skos-reference/skos.html#closeMatch

[10] https://www.w3.org/2009/08/skos-reference/skos.html#narrowMatch

- URIs represent one unique entity composed of a prefix, a time stamp and a suffix.
- The prefix contains the transfer protocol (HTTP), the web domain (data.novartis.net) and the resource (r1), which serves as URI syntax versioning, business domains and sub-domain to which the entity belongs and the system-of-record creating the entity.
- The prefix is strongly governed and defined by business and governance experts. The timestamp is either a version number or a date.
- The suffix indicates the entity type, standards that apply and parent-child relationship. Business experts define URI suffixes.
- This syntax is human and machine-readable. It is both a unique and unambiguous identifier for all entities in the KG and a pointer (indicator) to the entity's origin. It describes the business origin (business unit and/ or domain) and technical origin (system of record).

| PREFIX | | | TIME STAMP | SUFFIX | | |
|---|---|---|---|---|---|---|
| http://data.novartis.net/r1/business_domain/business_subdomain/system_of_record/time_stamp/type/standard/entity_n/entity_n-1/entity_n-2/... | | | | | | |
| **The URI prefix is composed by** | **Mandatory** | **Defined by** | **The time stamp is** | **The URI suffix is composed by** | **Mandatory** | **Defined by** |
| The domain for enterprise data | Yes | Enterprise DG | a version number | Type of data it represents | No | Enterprise DG, upper ontology |
| The release version for the URI syntax (r1) | Yes | Enterprise DG | or | Any standard applied to this data | No | Enterprise DG, upper ontology |
| Business domains and subdomains | Yes | Enterprise DG and authoritative source system(s) and person(s) owning the design of the organization structure | a date in ISO format | One parent data entity | Yes | DG and business experts |
| System of record | Yes | Enterprise DG and authoritative source system(s) and person(s) owning the inventory of systems/applications in the organization | defined by DG and business experts | Many child data entity | Yes | DG and business experts |

***Figure 2: Enterprise Uniform Resource Identifier (URI) syntaxe.***

## 5.3. Enterprise Resources

As a result of the above-described activities, unique, cleaned and linked (meta)data representing the enterprise business is integrated into the KG (Figure 3). The latter serves as a library to expose enterprise resources (artefacts) supporting DG4.0 and other goals such as communication, education, strategic advisory and the generation of new insights. The knowledge contained in the KG does not exist anywhere else in the company. This section lists all enterprise resources exposed from the KG as listed in the third column of Table 1.

As shown in Figure 3, the content of the KG comes from three internal sources (humans, documents and machines/ data) and external standards (blue boxes). At its core, there is a business glossary made of terms with many definitions (as per their business facets) linked to their corresponding sources and any other complementary information improving their meaning. Once defined, these terms serve as names for concepts, domains, inventory items, FAIRification or governance annotation, master or reference data or objects/attributes in data models (See previous section, “Knowledge Integration”). Our enterprise upper ontology (including the DG ontology) enables the description and prescription of governance principles to apply to 1) enterprise data assets and their content (object, attributes) and 2) metadata used to annotate these data assets (the full content of the KG). Importantly, the KG does not contain transactional and/or sensitive data but points to its: the KG exposes URLs where to find data assets in their source operating system, data warehouses, or lakes (yellow box). The content of our KG is transparently exposed in an intranet portal, and its content is accessible via a SPARQL endpoints API (orange box).

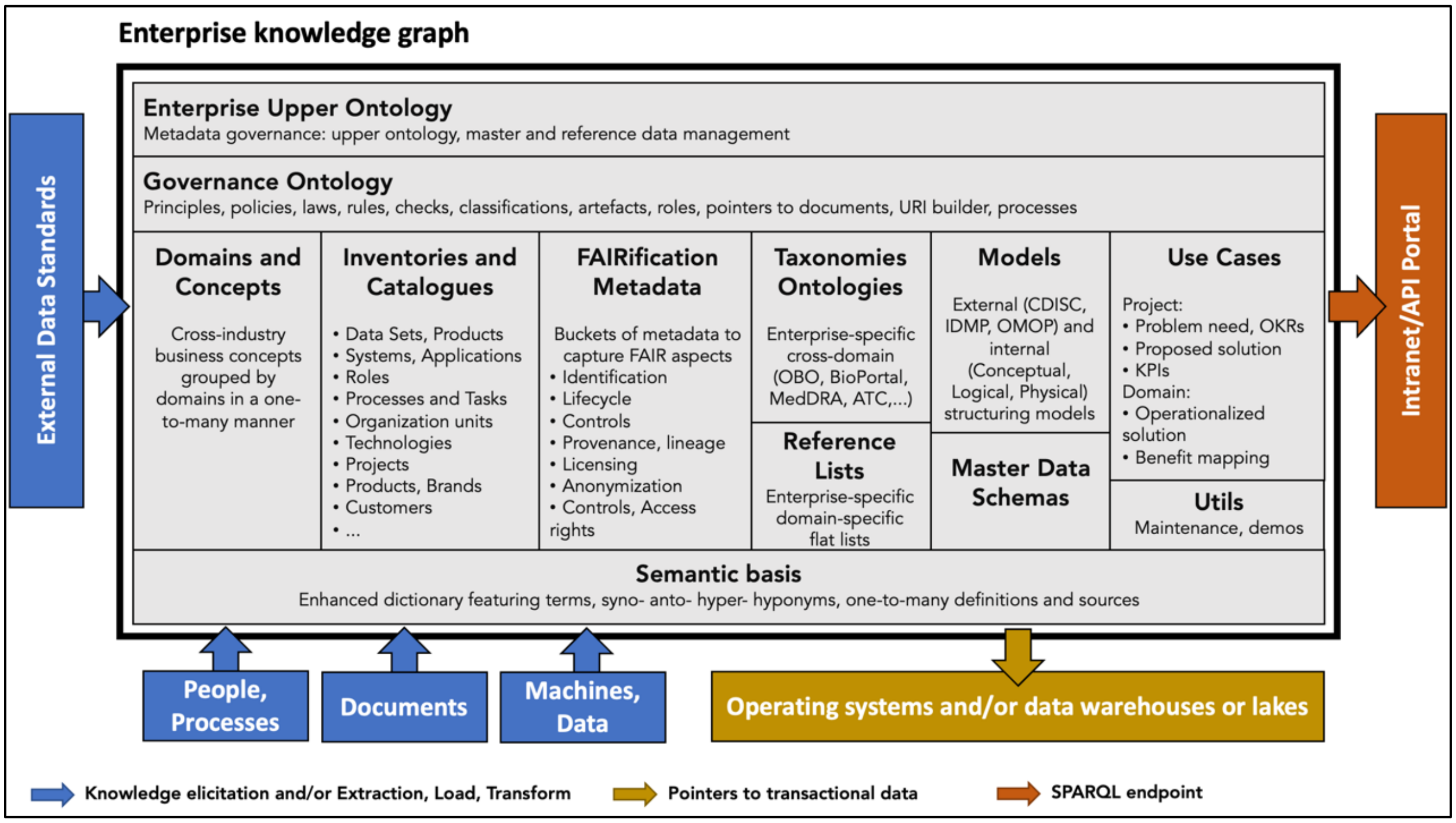

*Figure 3: Overarching content and architecture of our enterprise knowledge graph (KG).*

**The enterprise upper ontology (UO)** is made of public ontologies (such as Dublin Core, PORV-O, DCAT, P-PLAN, BPMN, FIBO...) coming from open sources (DOLCE, OBO, BioPortal) and enterprise-specific ontologies (IDMP) linked together using semantic enterprise architecture concepts (TOGAF, Zackman framework). It enables a consistent knowledge collection, maintains cohesion in the resulting KG and allows optimal knowledge retrieval via SPARQL queries. The UO contains a semantic version of business architecture standards that serves as the glue to link any piece of knowledge we collect across the organization and beyond (industry).

**The enterprise business glossary** combines internal and external business terms collected from knowledge elicitation (See section Knowledge Elicitation) or existing dictionary-glossary-thesaurus. We use the SKOS ontology to classify terms as "Concept" class[11], add preferred label[12], several alternative labels (synonyms)[13] and as many definitions as we find[14].

**Business domains** encompass one or many business concepts. When concepts are associated with EA entities (such as data sets, owners, governance, systems, transformation logic/algorithms and metadata), they can be treated as data products as defined by the data mesh theory (106). The KG exposes clearly how domains, concepts and products relate together; this is key to improving enterprise productivity along three axes:

1. This supports educational purposes as people realise that colleagues use the same concepts differently for other business cases (e.g. concepts like "customer", "product", and "market" are understood and used differently across the organization). The KG highlights the different understandings of the same (or similar) concepts across the end-to-end business process and reconciles them if needed.

---

[11] https://www.w3.org/2009/08/skos-reference/skos.html#Concept

[12] https://www.w3.org/2009/08/skos-reference/skos.html#prefLabel

[13] https://www.w3.org/2009/08/skos-reference/skos.html#altLabel

[14] https://www.w3.org/TR/2008/WD-skos-reference-20080829/skos.html#definition

2. Such business empathy is critical to improving purpose-driven or product-driven ways of working: it discloses how an employee uses upstream (digital) products to produce her/his product that will be used downstream.
3. It serves governance purposes to assign responsibilities at different level (business domains > concept > data set > object > attribute). It removes ambiguities at domain and data product interfaces and clarifies DG roles and responsibilities.

**Inventorying, cataloging and Searching:** Upon business analysis, we collect standard information about datasets that projects/teams use. This information revolves around dataset provenance (source), usage condition (license), and internalisation process (how they collect or obtain this data). A narrative (non-technical) description of data set content, the different roles involved in dataset onboarding, and the essential privacy and confidentiality relevance of dataset content complete the analysis. When organised and displayed consistently in the KG, this information serves to inventory source/raw datasets and data sources. Datasets are tagged with business concepts they relate to according to our overarching enterprise conceptual model (business domains and concepts described in the previous paragraph).

We inventory other business assets we encounter during business analysis, such as applications/systems, processes, roles, organisational units, projects, products, brands, customers, etc., as defined by EA building blocks. We created specific UO relationships to perform inter- and intra-inventory mapping. For instance, we link data sets to IT systems/applications and CRUDing[1] actions, or we link data sets to organisational units using them as part of projects. Catalogues are more customer-oriented and list products relevant to end-users rather than by the source. Catalogue items result from a transformation process, adding value to raw data. For instance, a catalogue of data products features newly created datasets containing clean and linked data from different sources; besides, data products have clear ownership and are fit-for-purpose to fulfil specific business needs or customer needs.

In conclusion, our KG serves as a knowledge base to link every piece of knowledge we collect from humans, documents or machines/data by following a structured knowledge elicitation process. As highlighted in Figure 3, subsets of the KG serve as content to populate several enterprise artefacts, such as:

1. **A business glossary**: business terms, synonyms, definitions and definition source (human, document or machine/data)
2. **A map of business concepts, domains** they belong to, actors managing them **and data products** they contribute to
3. **A data dictionary**: technical description of data sets and their content (link to data semantic data models)
4. **Information models** featuring concepts, objects, attributes, classes, and properties that contain; models and entities link to their definition, source and synonyms in the business glossary; concepts are linked to business domains
5. **A data catalog**: a single source of reference for data asset
6. **Catalogs of EA entities** linked to data assets (systems, processes, roles, organizations…)
7. **A catalog of master and reference data** annotated with FAIR tags and exposed in standard format for reuse across the organization
8. **A catalogue of FAIRified standards** and **mapping across standards** (internal or external list, taxonomies or ontologies)
9. **An enterprise ontology** including a **governance ontology**
10. **Multiple metrics, use case and examples** explaining the added value of our KG, its content and performances

# 6. Technical Implementation

While the Introduction and Background Information sections of this paper exposes the evolution of DG and corresponding inefficiencies, the Section 3 Knowledge Modeling and Future of Data Governance proposes a metadata model, a query language describing data assets and their business context is a prerequisite for an agile, fit-for-purpose and automated DG4.0. This section describes the technical underlying aspects of our technology stack, how it enables knowledge elicitation and integration, and the publication of enterprise artefacts listed in Figure 3 and in the conclusive paragraph of the previous section. Based on semantic web principles, our custom-developed platform includes technologies recommended by the Semantic Web Consortium (26). This section describes the technical aspects of knowledge ETL[15] or ELT[16] and knowledge access/retrieval. We provide additional details in the Supplementary Technical Information at the end of this paper.

The technical implementation that supports the execution of the proposed knowledge management methodology (Figure 1) is mainly requested for knowledge capture (step 1) and integration (step 2). This implementation is overviewed in Figure 4 and it details as follows:

1. During the business analysis, competency questions (problem statement), what success looks like, and any relevant information about business architecture entities (as defined by TOGAF) are identified.
2. Upon the knowledge elicitation step, information is captured digitally from all different types of sources and integrated.
    a. Information from humans is captured within templated Excel spreadsheets, and information contained within documents and machines is collected within data frames.
    b. Named entities are extracted from all captured data and benchmarked against our standard reference data, taxonomies, ontologies, and enterprise upper ontology (managed in Protégé). Especially for the corpus of documents, NLP techniques are used to extract named entities using a commercial semantic platform.
    c. Retrieved entities are compared with our business glossary benchmark, which may result in the glossary update if needed. Data sets and databases are profiled using custom Python code. Retrieved entities from table content and column are converted into RDF and compared with the KG, allowing them to be identified, tagged and disambiguated. Suggestions for similar or identical entities already existing in the graph are highlighted if needed for data unification.

[15] ETL: Extract, Transform, Load

[16] ELT: Extract, Load, Transform

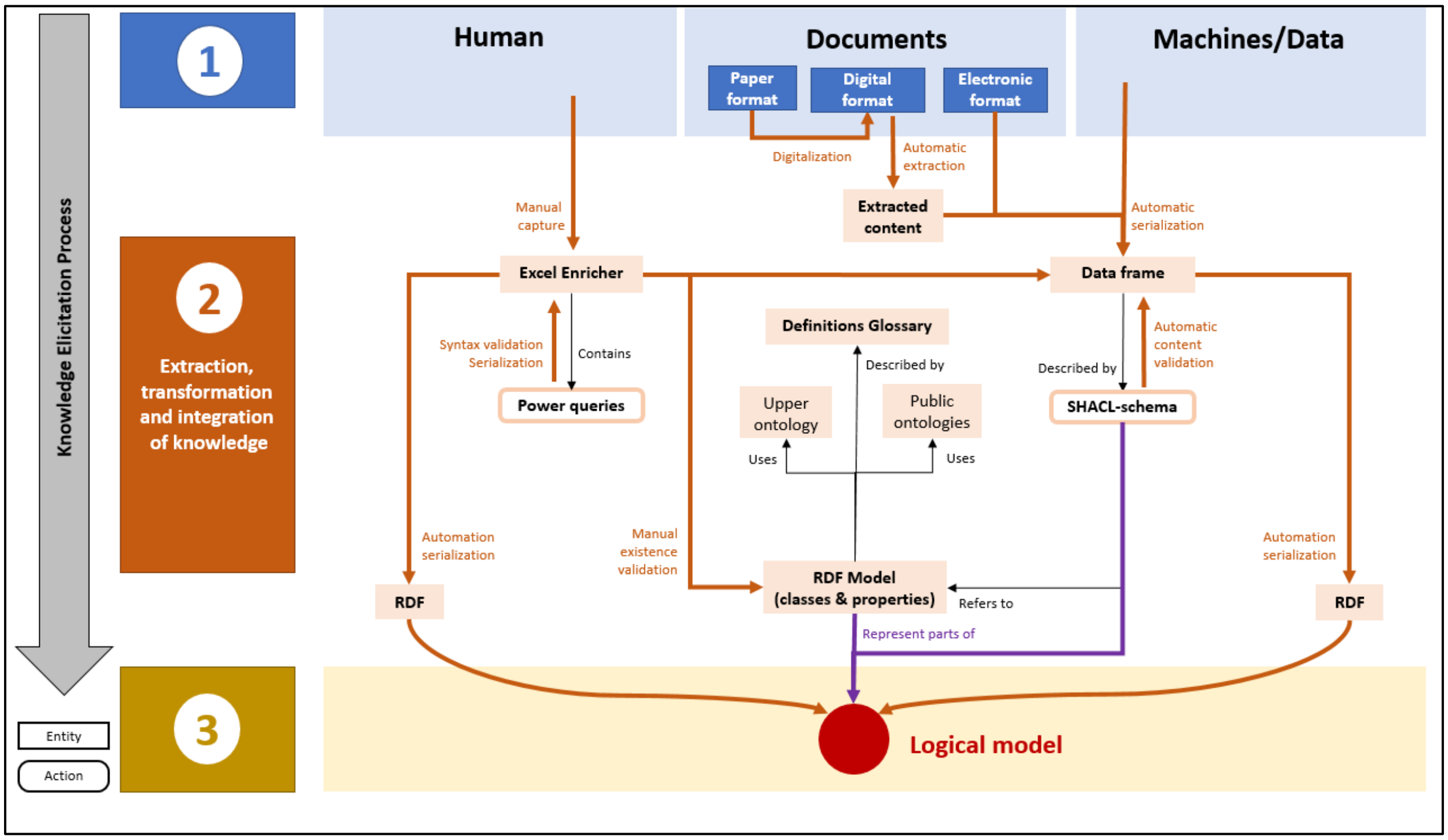

*Figure 4: Overview of the technical implementation that supports the knowledge management method.*

## 6.1. Source data capture, transformation, and validation

As described in the methodology section, relevant knowledge is presented in disparate formats, such as verbal communications (business discussion), documents (digitalised and non-digitalised), or systems and datasets. To ease knowledge capture and lower the barrier to adopting our solution, we use the world's most common data management tool: Microsoft® Excel. We standardized the knowledge capture and integration activities using a set of user-friendly Excel Enricher templates specialised to capture knowledge from each source (Figure 4):

- To be filled manually by humans upon discussions
- Using NLP to extract concepts from documents (not in the scope of this paper)
- For processing databases or tabular data sets in general datasets (data streaming is not in the scope of this paper)

### 6.1.1. Excel Enrichers for business discussions

Upon our business analyses, we realised that tacit knowledge about business "know-how" is complex to collect and thus poorly documented. We tried to develop user-friendly Excel templates (see Supplementary Information, section Framework Technology: supporting applications) so-called "Enrichers" that assist business experts in framing and capturing their knowledge using predefined phrasing: Excel cells are arranged to create simple sentences (subject, verb, object) that use standard upper ontology class and relationship to generate RDF-triples (Subject-Predicate-Object). These spreadsheets were designed to guide users so that they can reuse existing concepts when formalising their knowledge.

Excel Enricher templates contain three templated tabs to be filled out by business users with or without the help of data stewards. They are asked to capture the origin of information (for URI creation and lineage documentation) and generate a list of concepts commonly used in their field of expertise. Excel templates also facilitate the contextualisation of this knowledge. A set of automated quality steps ensures a correct URI syntax, unique preferred labels (per template), and proper concept definitions. A sequence of predefined queries (Figure 4, "Humans") generates a TTL file as output.

### 6.1.2. Knowledge extraction for documents

Extracting information within documents is a multi-step process to extract and categorise retrieved entities as per classes and relationships from our upper ontologies (Figure 4, "Documents"). We will not describe here common Natural Language Processing (107) steps. We will focus on how documents are annotated with retrieved entities.

Typically, digital documents in a PDF or Microsoft® Word format or .txt format are converted into electronic XHTML. Physical documents, such as paper format documents or scanned copies, require an additional pre-processing step of Optical Character Recognition (108) before content extraction. Resulting XHTML documents are easily visualised in any web browser. The XHTML format enables the direct application of text mining techniques to extract structural units of the document (such as headings, sections, titles, figures, and tables...), concepts from the bulk text and document metadata (authors, creation dates, version...). Using NLP solutions from a professional vendor, the following annotations were inserted inside the XHTML documents using RDFa (109).

- Text structures were identified and annotated with DoCO ontology (110)
- Publicly known biomedical concepts from OpenEHR (111) were extracted from bulk text
- Document metadata was identified and annotated using the Dublin Core ontology (100)

To increase the relevancy of retrieved OpenEHR concepts, we linked them with DoCO-identified sections in which they appear and perform basic frequency analyses. This activity enabled us to contextualise extracted concepts per each similar section across a document corpus and identify patterns (frequency concepts associations within a similar document section).

### 6.1.3. Semantic ETL pipeline for systems and datasets

During business discussions, we identify important data assets for a given project. Often, we need to ingest and transform the whole (or parts) of this data set as it contains relevant machine/data knowledge to our use case. These data sets come in various structured or semi-structured formats. To ETL this data at scale, we developed programmatic methods as an alternative to the Excel Enrichers. A collection of Python or Java codes is bundled into the package at the disposal of data engineers. The latter can generate RDF-triple at scale from large structured data files such as database extracts (See Supplementary Information section).

As detailed in the Knowledge Elicitation section, we link these data sets to their systems-of-record and ETL pipelines. Whenever available, data models describing the structure and content of these data sets (entity-relationship models or class diagrams at logical or physical levels) are converted or re-modelled in RDF format, loaded in the KG and linked to data set URIs.

Our semantic ETL pipeline starts with normalising the format of structured and semi-structured data sets. We apply matching algorithms (similar to OpenRefine (92)(93)(103)) that perform content harmonisation (as per upper ontology entities and reference data) and structure validation per predefined SHACL shapes. This step is crucial to obtain a logical representation (a model) of data sets that seamlessly integrates the KG and links to existing DG concepts (Figure 4**,** "Machines/Data").

All details associated with the semantic transformations (such as code, language, execution environment) are recorded in the KG and associated with the original input dataset and the output semantic model.

## 6.2. Knowledge Retrieval and Access

There are two ways to leverage the KG: visualise (subsets of) it as an intranet web page or access its content via SPARQL queries. The queries can be encoded in an API to provide predefined content or freely typed in a web client interface to the graph.

### 6.2.1 The DGSS portal

Subsets of KG are exposed in our Data Governance Semantic Suite (DGSS) intranet portal. The whole graph is too complex; there is no point in exposing it to end users. As illustrated in Supplementary Information in Figure S1a, DGSS enables a quick and easy exposure of all enterprise resources (listed in Figure 3 and conclusion of the Section 5.3 Enterprise resources) and other custom representations of businesses. One can easily retrieve any KG entities thanks to the embedded search engine, which is optimised for indexing RDF triples. Search results are tabular; except for literals, any KG entity can be clicked and displayed in a graphical form.

Each page in DGSS is built to satisfy specific business needs, as expressed during our Knowledge Elicitation process. Each page shows a subset of the KG as linked and cleaned data to answer specific business needs or competency questions (QCs). For governance purposes, specific DGSS expose the following pages as exemplified in Figure 3: maps of business domains, concepts and data products, the business glossary, catalogues of data assets, systems, processes, roles, data dictionaries with FAIR principles and corresponding (logical) models, mapping between logical and physical models, …

It is important to note that DGSS does not expose transactional data or sensitive data. The entire content of the KG is risk-classified as "business use". For instance, the portal does not contain instantiation of clinical or commercial data that may be of private or confidential nature. Instead, DGSS expose pointers to this data in their respective system-of-record. As a result, access to DGSS content is not restricted and all pages are transparently available to all employees.

### 6.2.2 The DGSS FAIR API portal

The DGSS FAIR API portal (Supplementary Information in Figure S1b) is a catalogue of SPARQL queries that retrieve data from the KG. Each query (represented in the KG) is annotated with FAIR and upper ontology concepts as per the KG content it gets. FAIR and upper ontology apply to the data payload that the SPARQL query retrieves. Because the retrieved data is richly annotated with metadata (domain, owner, DG roles, provenance/lineage), we can consider it a data product (106). Each SPARQL query can be regarded as a simple API to get KG content that answers the specific business needs or competency questions (QCs) expressed during our Knowledge Elicitation process.

# 7. Data Governance 4.0 Applied to a Unified Clinical Data Model

Good data starts with Great Governance (13). Better data continues with automated and agile governance adjusted to business needs. These are fundamental principles underpinning DG4.0. To achieve DG4.0 goals, we propose conducting structured Knowledge Elicitation to feed and maintain a rich FAIR knowledge repository materialised as an enterprise Knowledge Graph (KG). The previous section explains how to access, explore and retrieve knowledge from the KG via an intranet portal (DGSS, see Supplementary Information in Figure S1a) or FAIR API portal (Supplementary Information in Figure S1b). In this section, we provide a concrete example of how DG4.0 principles support creating, maintaining, and governance a model that unifies 20 years of legacy clinical study data in one target model.

## 7.1. Unified Clinical Model

While previous sections explain why and how we built an enterprise KG that enables agile and automated DG4.0, this section illustrates our claims by focusing on one concrete example to unify (pool under a common schema) 20 years of clinical research (3'500 clinical studies, 900'000 subjects, over 3 million source variables categorised in 100'000 domains). Therefore, we created a Unified Clinical Model (UCM) to map source variables to 4000 target variables belonging to 50 domains.

This section exemplifies how the UCM supports data harmonisation across three therapeutic areas. Regardless of therapeutic areas, clinical data is structured according to one of the three standards for clinical data: NovDD (legacy format), GSK (coming from the GSK oncology pipeline acquired by Novartis in 2014-2015)(112) and the current Novartis Clinical Data Standard (NCDS).

The UCM is FAIR by design leveraging URIs and rich meta-metadata[17] to describe itself. On top of being a FAIR transformation recipe to map source variables to target variables, this model is also an enabler for DG4.0, leveraging and integrated with all activities and resources listed in Table 1 such as:

- provenance/lineage of clinical data
- conformance to external data standards (CDISC)
- data quality and anonymisation checks
- linking to enterprise/business architecture resources
- enterprise data asset inventory
- insight generation for business process improvement

## 7.2. UCM Integration to Data Governance 4.0 framework

Figure 5 depicts the UCM and how it integrates/links with resources from our enterprise DG4.0 framework. The yellow triangle represents the UCM linked to other entities (blue and orange boxes) from our KG. The basis of the triangle provides numbers about source data sets (clinical study level), while the top of the triangle shows numbers about the target data set (target model). The mapping of sources to target involves around 25'000 transformations; all described semantically with a group of predefined metadata that provides meaning to transformations.

[17] We would like to point out the fact that master-, meta-, reference- or transactional data (data typing or categorisation in general) is always relative to the data user. The UCM being metadata to clinical data users, data describing the UCM can be considered metadata to metadata for clinical data users.

Source entities, target entities and transformations are richly and consistently described (annotated) using entities from our enterprise's upper ontology. Blue boxes represent data/information about the business context and usage of the UCM. In contrast, orange boxes represent data/information about the business process in which the UCM is used: the UCM is the pivotal point of a process starting with writing the study protocol to the publication of results in the clinical study report.

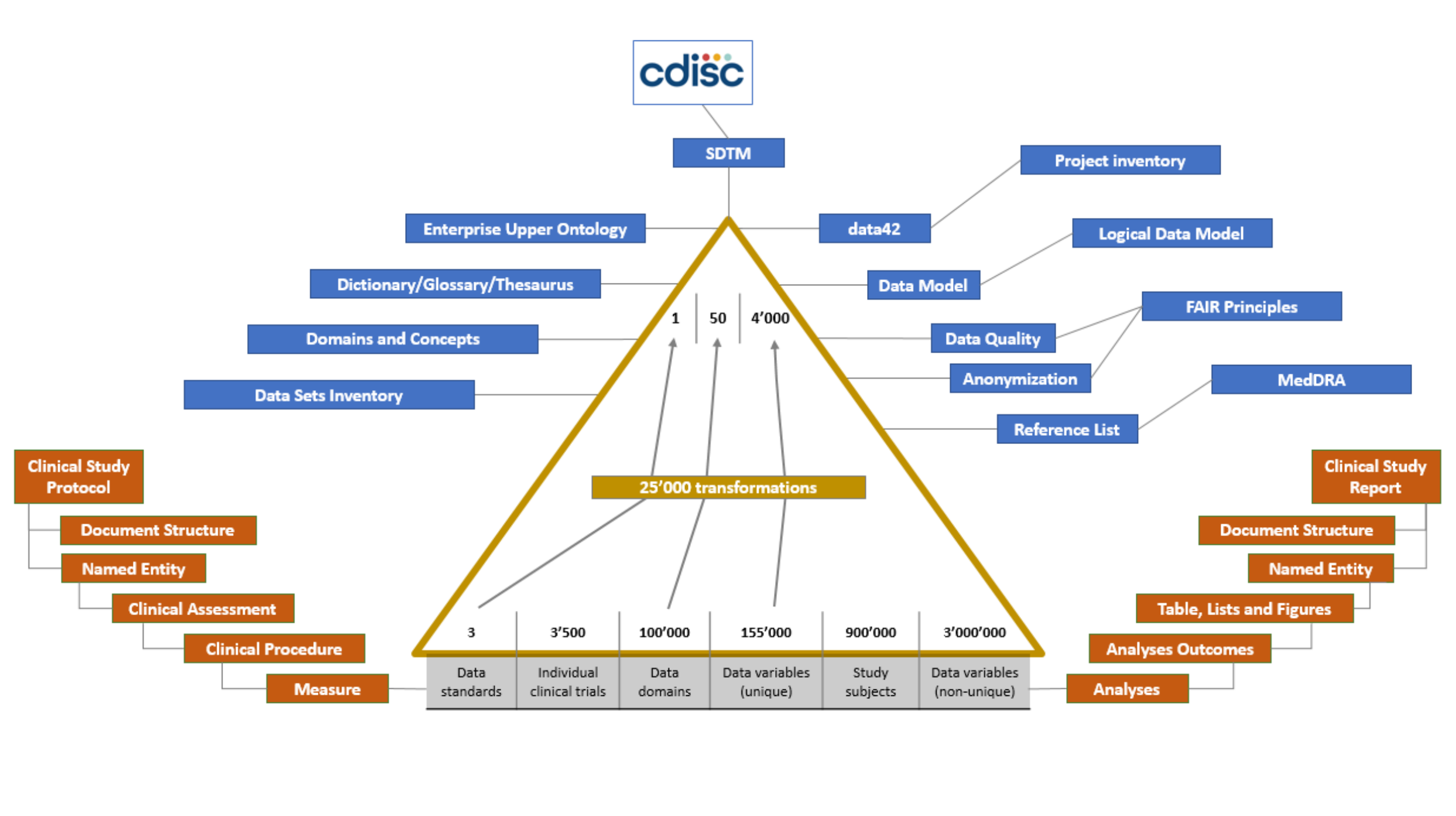

*Figure 5: The Unified Clinical data Model (UCM) in context*

First comes the integration with the enterprise enhanced business glossary: clinical domains and variable are defined by one or more definitions listed in the glossary along with alternate and/or hidden labels (if it applies).

Secondly, the UCM integrates with the enterprise conceptual data model. The UCM mainly covers "Preclinical Research" and "Clinical Development" enterprise business domains, which are different from "CDISC" clinical domains. Figure 5 shows how the UCM links to DG4.0 artefacts (blue boxes on the left-hand side); all these artefacts are linked in the background as they all come from the KG. Therefore, UCM variables are linked to CDISC domains that link further to enterprise/industry concepts grouped by business domains. We also link to enterprise master data (such as "Clinical Study", "Compound", "Medical Indication", Mechanism of Action", "Clinical Study Subject", etc.) that are all characterised using govern attributes/properties. Master data attributes/properties are instantiated using controlled terms or values from internal or external standard terminologies (e.g. CDISC code lists, MedDRA, MeSH, WHO ICD, etc.).

Thirdly, the UCM links to enterprise/business architecture artefacts (blue boxes on the left side of Figure 5), such as Projects[18], data models, data quality rules and checks and other standards.

[18] Either so-called "Molecular projects" contributing to the R&D of one molecule in our portfolio or "Non-molecular project" not directly related to the development of a drug candidate.

## 7.3. Source Variable Provenance and Versioning

All source variables have a governed URI linked to a preferred label, alternative labels and one or more definition(s) (see The Enterprise Business Glossary in the previous section). As described in section Metadata governance, URIs are a unique and unambiguous identifier for clinical data and a human- and machine-readable indicator of the source from which clinical variables or domains are coming. The URI syntax for a UCM variable is built as follows:

> *URI prefix: development/clinical/globalmetadata/*
>
> *URI time stamp: v1/*
>
> *URI suffix: fullyqualifiedelement/DR_AA_AACAT*

Carrying URIs throughout the entire lifecycle of a clinical data set ensures unicity, explicit characterization (URI suffix) and provenance of data (URI prefix) as explained in many use cases listed in Table 1. In addition, variables have "fully qualified" preferred label that provide all relevant information (standard, stage, domain they belong to and variable name) to identify any variable unambiguously.

Although it is not a W3C recommendation (104), we choose to insert a time stamp in the URI syntax. Our knowledge elicitation process mostly relies on templated spreadsheets which automatically capture authorship metadata including the day date. When a data expert uses the spreadsheet to capture/update information, it automatically generates the proper URI including a date in ISO format. While this approach guarantees proper batch URI creation/update it does not allow individual URI editing.

## 7.4. Source to Target Many-to-One Mapping

Source data is available in different data standards (NovDD, GSK and NCDS). These standards contain different entities (standard, domain, variables) and different reference lists (controlled vocabularies). Although they had the merit to exist, legacy standards were less enforced than CDISC-based standards. As a result, standards were considered more as guidance and inconsistently applied in old clinical studies. This resulted in clinical study data claiming compliance to the same standard but still difficult to reconcile.

Predefined metadata sets characterise each source variable, domain and transformation. Figure 6 is a screenshot of our KG exposing requirements to map three source variables (capturing the gender of clinical study subjects) to one target variable. This figure provides the recipe for the mapping of three different variables representing the sex or gender of study subjects collected as per three different data standards. Internal clinical data standards (up left-hand corner, GSK, NovDD and NCDS) decompose into domains ([DEMO], [DMG], [DM]) that contain variables (GSK.DEMO.SEX, NOVDD.DMG.SEX1C, DM.SEX). GSK and NovDD variables need to be transformed via several operations (or derivations, Deriv #) to eventually become the DR.DM.SEX target variable. The DR.DM.SEX variable becomes the standard CDISC SDTM variable SD.DM.SEX via a "COPY_ELEMENT" operation. This variable is linked to a reference list called "SEX" that contains the permissible value the variable can take, namely, "Male", "Female", "Undifferentiated" or "Unknown".

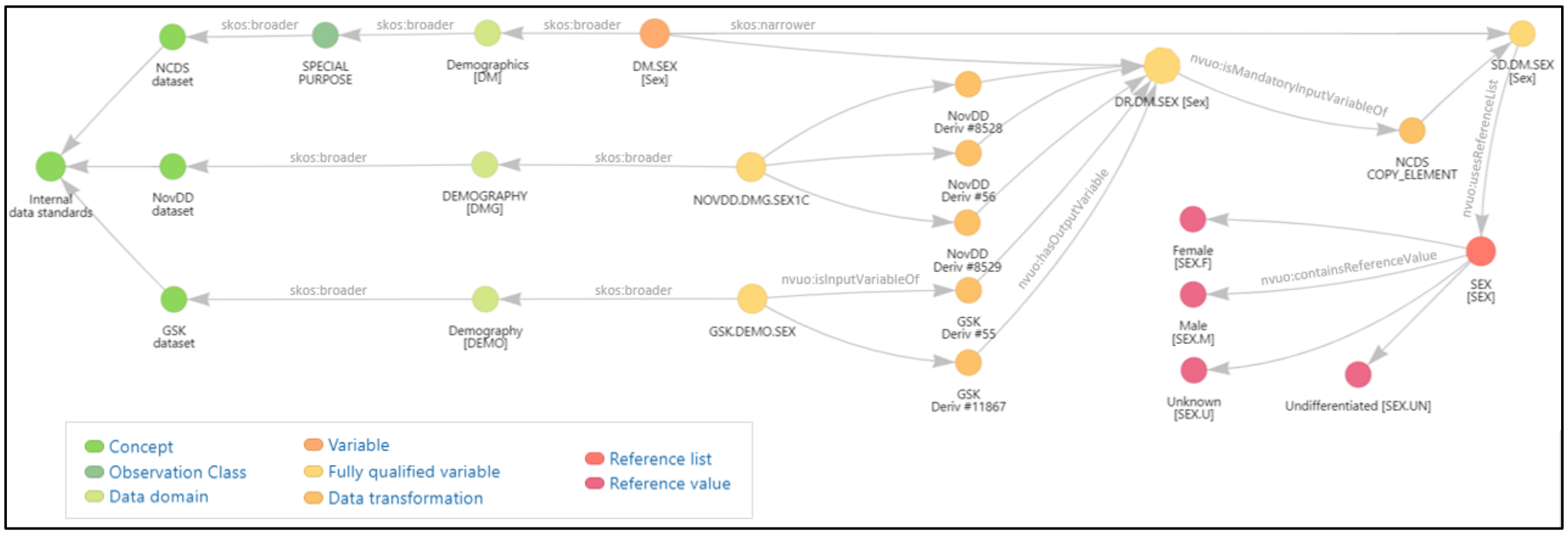

*Figure 6: Example of sources to target mapping of the subject gender variables.*

## 7.5. Variable Lineage across Transformations

We define as lineage the documentation describing all transformations that one source variable undergoes to map to one target variable. We capture both a business- and technology-lineage manner. The business lineage links the different stages variables take over time. In contrast, the technical lineage is more transformation-centric and documents whether a variable is an input or an output, to or from one or many transformations.

Figure 7 illustrates the entire lineage from many source variables to a unique target variable across multiple transformations, which is documented in the KG. This figure focuses on the DR.DM.SEX variable. The bottom part of the figure shows technical mapping, i.e. transformations/derivations providing DR.DM.SEX as output and other transformations/derivations taking DR.DM.SEX as input. The middle part shows variables upstream (left to DR.DM.SEX) and downstream (right to DR.DM.SEX) in the process. The upper part shows how the DR.DM.SEX variable maps to CDISC SDTM standard. The right panel shows how the DR.DM.SEX variable is linked to the corresponding reference list (SEX) and controlled vocabulary composing it ("Male", "Female", "Undifferentiated", "Unknown").

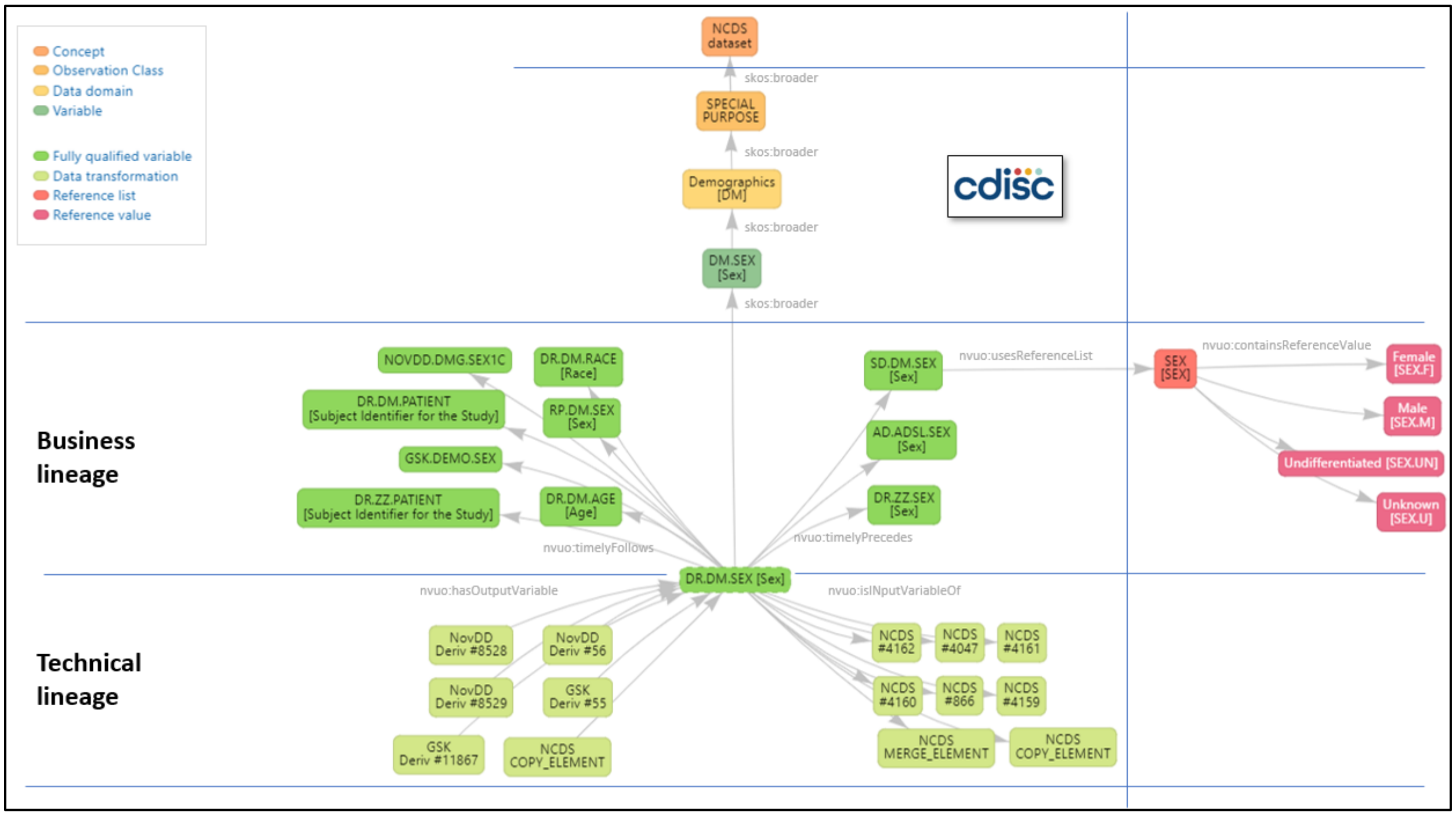

*Figure 7: Business and technical lineages of variables in the UCM.*

## 7.6. Synching the UCM with Data Standards

The current Novartis Clinical Data Standard (NCDS) is based on the CDISC SDTM standard (113). We feed from CDISC public APIs, convert the information into a semantic format, and use an ontology management tool to integrate it into the KG. Data is updated regularly using a semi-automated process (see section External Data Standards Management).

Per health authority requirements, CDISC standard vocabularies (code lists) must apply when instantiating variables in submitted datasets. Similarly, other standards might apply as per requirements from another authoritative actor. The UCM links variables to all code lists that apply for given circumstances. These lists often come from external standards via APIs. As described in the Knowledge Integration section, we ETL these standards in a (semi)automated way and perform an inter- and intra-list mapping using fuzzy logic. Then, we transparently expose variables, code lists and the logic that applies to bind them.

## 7.7. Data Quality and Anonymisation

We expose the UCM as a data product to be consumed by scientists (customers) who need to combine clinical data on demand at high speed and large scale/volume. Keeping the UCM of high quality is critical to customer satisfaction.

To ensure UCM quality, we link UCM entities (standard, domain, variables) to KG nodes representing quality principles, rules or checks to consider for quality assurance. These concepts ensuring KG quality are part of our DG ontology (see section Data Governance). The DG ontology is an integrative part of our upper ontology that serves as a prescriptive source of requirements that UCM users shall apply to ensure quality (meta)data.

Figure 8 shows how quality checks are derived from high-level data quality principles cascading to UCM variables. A general governance principle, “ALCOA+”, covers several data quality issues. One of them is data “Completeness” and relates to specific rules. One rule consists of checking whether data values are missing. The rule is instantiated in the KG as an action “Missing value check” that links to variables for which this check applies. One of these variables is called AE.AEDECOD, which belongs to the CDISC SDTM “Adverse event” domain. In our DG4.0 framework, we have generalised and consistently applied such a graph pattern to high-level principles (data quality or other governance principles), rules, and data objects or attributes they apply to. When published as prescriptive reports, these principles list actions to take to govern (including curative actions) the corporate data landscape.

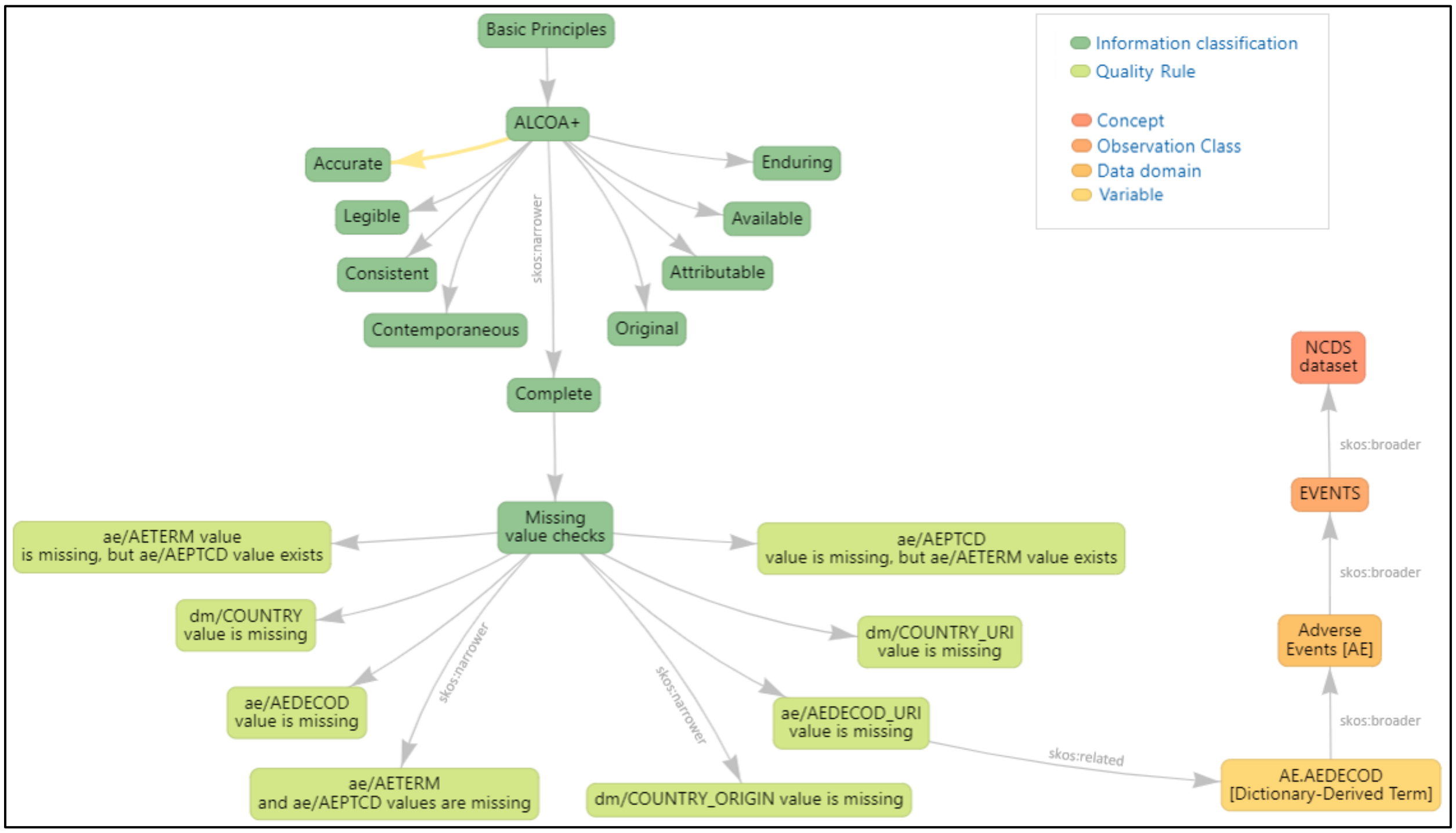

***Figure 8: Linking governance principles to data to which it applies.***

The same approach supports data anonymisation. Anonymisation principles, rules and checks are stored in the KG/ UCM and linked to privacy- or confidentiality-relevant variables. These rules can be the "suppression" or "noise addition" to numerical variables or the "offset" date variable. Anonymisation metadata in the KG specifies whether such modification shall apply in a generic or specific manner (for instance, only for a particular clinical study). Reporting on anonymisation metadata supports data reuse (R in FAIR); this metadata describes to what extent the raw clinical data was modified for anonymisation purposes.

Metadata is also required to measure the risk for de-anonymisation (i.e. chances that a subject is uniquely identified despite initial anonymisation precautions) whenever anonymised data is iteratively combined with new data. This is a significant risk to monitor since we consider anonymisation a dynamic transformation depending on several parameters, such as the use case, the user(s), and the successive analytical steps the data goes through (data lineage). Because our KG integrates these parameters, we can report on different combinations.

## 7.8. Linking UCM to Study Protocol for Business Process Improvement

Clinical data stored in our clinical study databases are collected via standard clinical procedures conducted by healthcare professionals as part of our clinical studies. Each clinical study protocol describes these procedures in detail as narrative text (unstructured data) or tables (semi-structured data). In the UCM, clinical procedures are defined in the ontology and organised as taxonomies; the latter are linked to measurements and values (UCM variables) they generate. These values are then analysed and represented as a table, a list of figures in clinical study reports.

As shown in the orange boxes of Figure 5, the UCM contains the metadata linking concepts in the study protocol (far left orange box) to clinical procedures, measurements, UCM variables, and so on down to the clinical study report. For instance, the “Safety evaluation” section of the protocol mentions in a “schedule of assessments” several “blood draw(s)” to measure the concentration of “liver enzymes”; this ASAT[1] and ALAT[2] concentration is captured with specific variables to be used in predefined analysis and represented in standard tables in the clinical study report. Propagating this (lineage) metadata at protocol writing time can significantly reduce the time required to produce downstream artefacts by pre-populating them (e.g., clinical database schema, analysis algorithms, tables, lists, and figures).

By linking end-to-end concepts from study protocols to concepts in the clinical study reports, we are able to first, support content reuse and semi-automatise artefact writing/creation and second, identify bottlenecks in the process and suggest improvements to shorten the time elapsed between clinical study protocol writing and the first visit of the first patient (114).

## 7.9 Capture and Standardisation of Adverse Event Date and Time

More complex examples can also be extracted using the semantic graph querying capabilities. Submitting a clinical study to health authorities requires submitting all data according to CDISC standards. To this aim, we need to map from the collection to the analysis format (Figure 9).

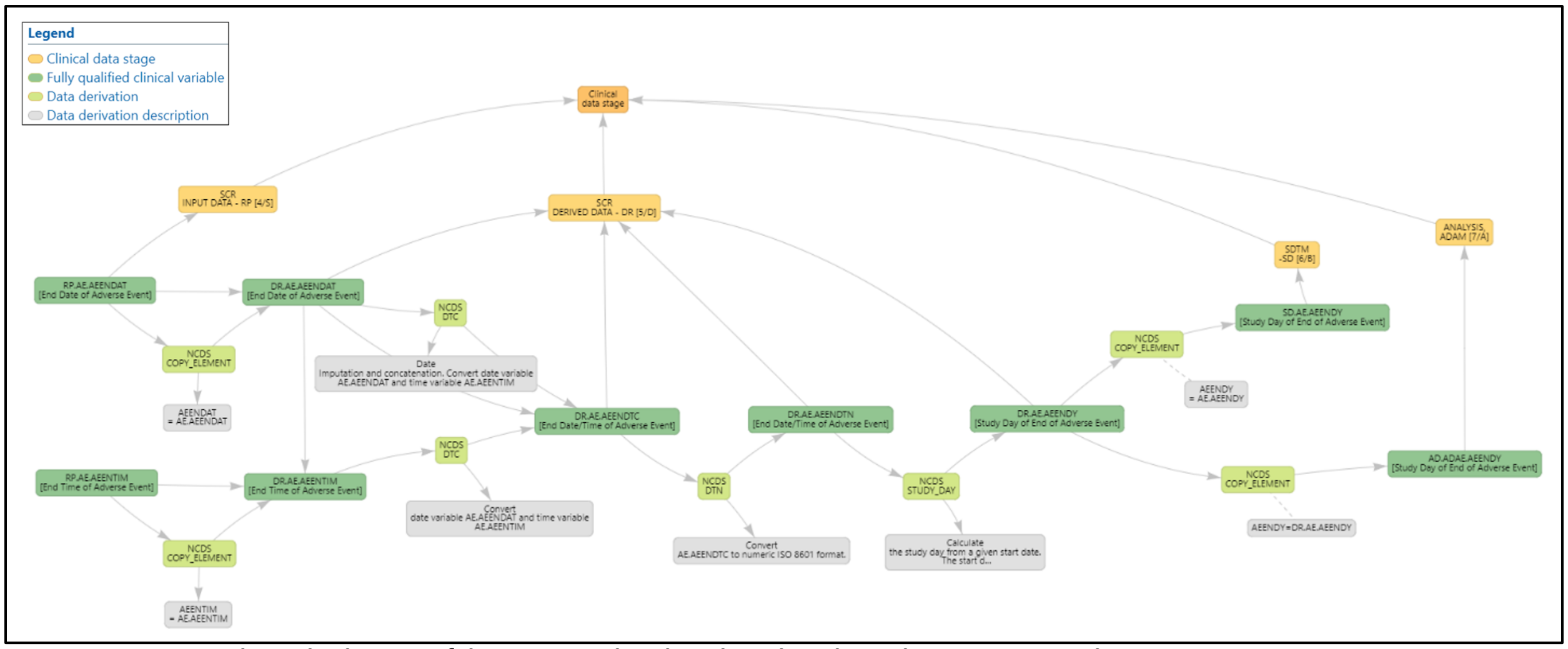

***Figure 9: Capture and standardisation of data points related to Clinical Study > Adverse Event > End Time & Date***

The collection format [SCR INPUT DATA – RP] is used by medical practitioners to provide data. This example shows two entries characterising “the end” of an adverse event: a Date variable [RP.AE.AEENDAT] as well as a Time variable [RP.AE.AEENTIM]. The first stage in which transformations are performed is called Derived [SCR DERIVED DATA – DR]. In this example, our objective is to create a variable containing the number of days from the start of the study to the end of an adverse event [AEENDY]. We avoid going back and forth between stages by bringing the two previously mentioned variables to the DR stage using a copy [NCDS COPY_ELEMENT]. A character “date & time” variable is created with the transformation DTC [NCDS DTC]. It consists of the combination of the date and the time variable. From this “Date” in a character format, it is necessary to convert it to a numeric value [DR.AE.AEENDTN].

From this numeric date and the date of the start of the study (not represented), we can perform a calculation of the study day [NCDS STUDY_DAY], thus creating our new "Study Day of end of Adverse Event" variable [DR.AE.AEENDY]. As per the submission requirement, there is a need to submit in the CDISC SDTM standard, hence the creation of the [SD.AE.AEENDY] variable, which is the SDTM Adverse Event variable. However, there is a last stage, when this variable is used for the analysis. Therefore, in graphs or other time points calculations, we need the variable to be presented in the CDISC ADaM standard [Analysis Data Model]. We now have the path of transformation from two collected variables of date and time to the study day variable in the analysis dataset.

### 7.9.1 Information accessibility

The information described above (Figure 10) was extracted using the power of the semantic graph querying capabilities (SPARQL). All use case requirements were accessed and extracted simultaneously to satisfy business needs.

More specifically, Figure 9 depicts, in a graph, the process of capture and standardisation of data points related to Clinical Study > Adverse Event > End Time & Date from the source format (INPUT DATA – RP, Data Stage 4) down to (SDTM – SD, Data Stage 6) and (ANALYSIS ADAM, Data Stage 7). Figure 10 shows the SPARQL query used to obtain all combinations of [Input Variable] <> [Derivation] <> [Output Variable] that relate to the "AE.AEEN" conceptual variable. "AE.AEEN" stands for "Adverse Event Adverse Event End" (date, time or datetime). Table 2 shows some variables (input and output to derivations) retrieved when using the SPARQL query described in Figure 10.

```
SELECT DISTINCT ?Input_Data_Stage ?Input_Variable ?Derivation ?Derivation_Rule ?
Output_Variable ?Output_Data_Stage
WHERE {
      # main search criteria
      FILTER(REGEX(?Conceptual_Output_Variable,"AE.AEEN")) .
      ?Output_Variable_URI skos:broader ?Conceptual_Output_Variable_URI .

      # search contraint
      ?Input_Variable_URI :isInputOf ?Derivation_URI .
      ?Derivation_URI :hasOutput ?Output_Variable_URI .
      ?Derivation_URI :transformationRule ?Derivation_Rule .
      ?Input_Variable_URI :dataStage ?Input_Data_Stage .
      ?Output_Variable_URI :dataStage ?Output_Data_Stage .

      # labeling
      ?Conceptual_Output_Variable_URI skos:prefLabel ?Conceptual_Output_Variable .
      ?Input_Variable_URI skos:prefLabel ?Input_Variable .
      ?Output_Variable_URI skos:prefLabel ?Output_Variable .
      ?Derivation_URI skos:prefLabel ?Derivation .
}
ORDER BY ?Input_Data_Stage ?Output_Data_Stage ?Input_Variable ?Output_Variable
```

***Figure 10: SPARQL query used to generate the entities described in*** Figure 9.

# 8. Data Governance 4.0 Framework Evaluation

The Introduction and Problem Statement sections describe legacy DG and associated paradoxes or challenges related to digital awareness and operational aspects. In this paper, we report our experience in addressing these issues by implementing an innovative Data Governance (DG) framework leveraging a knowledge graph (KG) based on semantic web principles. When searching for existing literature in commonly used databases[1], we found no article explaining how a semantic KG (RDF triple store) can support DG. This article lays the foundations for an agile, fit-for-purpose, and (semi)automated DG that we call Data Governance 4.0 (DG4.0) as per the fourth industrial revolution.

After providing some background information essential to DG4.0 basis (Background information), we explain in section 4 (Knowledge Modelling and the Future of DG) how knowledge models enable DG4.0. We emphasize how open standards such as semantic web principles are FAIR by-design and how they can be used to FAIRify enterprise assets. As defined by TOGAF or the Zackman framework, such enterprise assets enable a detailed description of the business architecture (context), embedding data assets subject to governance. In section 5 (Knowledge Management Methodology), we explain the knowledge elicitation method that feeds our KG and how we capture knowledge from humans (tacit knowledge, know-how), from documents (text-based unstructured data) and data/machines (structured data). We then focus on how this knowledge is integrated according to governance activities defined by the DMBOK2.0 and FAIR principles. Section 6 (Technical Implementation) focuses on the technical aspects of our framework. We describe the process and the technological stack we use to ETL or ELT knowledge from different sources and how we expose it further in the organization. Section 7 (DG4.0 applied to a Unified Clinical Model) illustrates our claims by providing several concrete examples of how we govern a model to unify clinical data sets acquired over 20 years of conducting clinical trials.

**The DG4.0 framework is technically easy to implement, cost-efficient and flexible**. Several open-source or commercial KG solutions [https://github.com/totogo/awesome-knowledge-graph] can be installed on-premises or in the cloud for a reasonable cost. It does not require significant upfront investments. KG content is retrieved with SPARQL queries using a web client application. Some investment is needed to design a user-friendly interface (see Supplementary Information in Figure S1a and S1b). Knowledge ingestion (inbound to the graph) and publication (outbound from the graph) are matters of ETL/ELT to or from W3C standard formats. Leveraging W3C standards regarding technology and data format contributes to lower DG costs (see Problem Statement Paradox 1) and supports local DG initiatives while showing how they contribute to the big picture (see Problem Statement Paradox 2). Because the graph's content is exposed transparently to the entire organization, it creates trust among stakeholders (Challenge 5) and increases end users' data literacy (Challenge 7). Also, using open standards ensures the KG content is not siloed in a proprietary format or process and can serve as a basis for building future solutions; it is a more sustainable and resilient solution than vendor-specific DG solutions (Challenge 10). Regarding overarching system architecture, implementing the DG4.0 framework is not disruptive and comes as an overarching read-only system that collects metadata and governance knowledge.

**The DG4.0 framework supports the traditional DG area**. As listed in Table 1, the DG4.0 framework enables the publication of business-friendly enterprise resources (column 3) that support all DG capabilities listed in the DMBOK2.0 (column 4), such as terminology management (lists such as business glossaries, taxonomies, ontologies), business/enterprise architecture (including data and process modelling), enterprise data integration and lineage, definition and mapping of master and reference data, data product management and (meta)data FAIRification (see list end of section 5.3 Enterprise Resources).

**The DG4.0 framework supports fit-for-purpose, agile and (semi)automated DG.** Describing data assets in their multi-faceted business context (describe using TOGAF, Zackman architecture bricks) enables one to understand their multiple business usages and adapt their governance accordingly. Specific governance principles, rules and checks are detailed in the KG and linked to the data assets they apply. This is a key prerequisite for fit-for-purpose governance. The latter can be adjusted based on changing business needs by modifying governance principles/rules/checks in the graph; DG reports from the graph will change accordingly and help DG actors (authoritative bodies, stewards, owners) to adapt the DG processes, including global and/or local activities. Regarding (semi)automation of DG, it is important to understand that the content of the KG is machine-readable. The graph serves as a reference library to feed DG requirements to the system/application; for instance, specific entities in the graph can configure APIs, hence embedding governance rules and checks into data flows outbound from the KG.

**The DG4.0 framework fosters digital transformation and awareness across the organization.** The KG links local (project-level) governance activities, roles, and systems to global (enterprise-level) vision, mission, objectives, and digital transformation guidelines. By transparently showing how local governance contribution articulates with global guidance, stakeholders understand how their input contributes to the big picture; this gives more sense to the extra effort they have to provide and increase engagement (see Problem Statement, Paradox 3).

**The DG4.0 requires new skills and heavy maintenance and business engagement while scaling-up.** Managing a KG and adopting semantic web principles is not easy. The learning curve is steep and counterintuitive to people who handle relational databases. The framework requires new skills and capabilities that an existing IT organization might not necessarily have. The maintenance of high-quality metadata in the KG is of paramount importance for DG4.0. Maintaining the KG content up-to-date and as rich as possible is a real challenge. From a business point of view, it requires engagement and continuous support, hence the importance of having a solid knowledge elicitation process. From an IT point of view, the effort is directly correlated to the maturity of the IT landscape and, more precisely, the ability of systems/applications to exchange data using W3C standard formats. At the core of semantic web principles are URIs. Very few commercial systems can leverage RDF URIs.

## 8.1. Contributions to literature and Implications for practice

DG4.0 enables (meta)data-driven governance. Although we developed it and applied it in the pharmaceutical/healthcare industry, DG4.0 principles hold for other industries. DG4.0 opens a new perspective regarding data management/governance maturity and might be taken into account to complement the CMMI Data Management Maturity Model (115).

DG4.0 enables the FAIRification of data assets. This might be a helpful framework to support the go-fair (34) and FAIR plus initiatives (116).

DG4.0 can support modern data management domains (117) and the definition of data products as part of a data mesh (86)(87)(118).

Govern data products are analysis-ready and can train AI models. It is now accepted that big data is not sufficient and that clean data (thanks to adequate governance) is the bottleneck for AI to deliver on promises (119).

# 9. Conclusion

Our journey to develop and implement this solution was not easy. Like any innovation, it faces inertia and interests of established working practices and parties. Potential difficulties for scaling up and maintaining content of good quality were often opposed to our approach. Although we acknowledge these challenges, they don't weigh much compared to the enormous benefits a DG4.0 framework brings. DG4.0 shifts the DG paradigm from a central, top-down, bureaucratic, restrictive and regulation-driven process toward a decentralised, community-led, digital process that transparently enables an organisation's digital transformation.

DG4.0 is only possible using a KG as a metadata lake. As metadata becomes big data, it becomes a wealth of knowledge to unify and tap into to drive business improvements via DG or other approaches to describe current business processes, prescriptive improvements, or predict risks. If we get the metadata right, we'll get the data systematically and longitudinally (over time). Moreover, a metadata KG based on open standards (W3C) is more resilient to changes than proprietary solutions. Compared to vendor-provided DG solutions, we think a DG4.0 framework is more sustainable and contributes less to increasing an organization's technical debt.

To conclude, we think organizations that can quickly plug and play new data sets on demand (irrespective of their size or format) will develop a significant competitive advantage. While plugging a dataset is a technical issue, playing the data is the hardest part. This is where a rich knowledge of data assets and their business context is instrumental. Integrate new data sources seamlessly into an existing data lake can only take place under two conditions: first that existing data sets are readily available with high-quality content (FAIR data), and second, there exists a rich knowledge source or model (FAIR metadata) documenting these data sets that one can query. Both conditions are satisfied using a Data Governance 4.0 framework.

# 10. Declaration section

## 10.1. Authors' contributions

CB initiated the project; conceptualisation, methodology and project administration were driven by CB and SM; CB did investigation and supervision of the project; BM, BJ, TM and MO implemented the technology and coded the integration and user interface; CB, PJ and CK performed data modelling.

Data curation, formal analysis and validation of the Unified Clinical data Model was done by CB, SM and AG; use cases were compiled by LH; GH conducted, managed and developed the code for named entity extraction from clinical study protocol; MO and TM developed the Semantic ETL methods in python.

CB, MO and SM wrote the manuscript, integrated individual contributions from each author and oversee the review.

## 10.2. Acknowledgements

Thanks to Hrvoje Mohoric to provide the adequate sponsoring and working environment that enabled us to innovate and propose new data-centric ways of working. Thanks to Andrea Splendiani and Ratnesh Sahay for a thorough review of this manuscript.

| Index | RDF use case | Captured and integrated knowledge | Enterprise resource it enables | Associated DMBOK2.0 data governance area |
|---|---|---|---|---|
| **1** | Business description | - Business needs, QCs<br>- Processes and roles<br>- Data source of truth (roles and systems)<br>- Data provenance<br>- Basic understanding about the privacy and confidentiality aspects of data<br>- Governance metadata such as steward identifier, date, base URI syntax and other optional notes | - Description of digital assets in their business context<br>- Catalog of competency questions and proposed answer as a subset of the enterprise KG in an intranet page<br>- Data sensitivity status and associated risks<br>- Display and justify governance principles based on risks<br>- Data governance reports for each data sets with contributing principles and roles | - Meta-data |
| **2** | Terminology management | - Business terms, definitions, synonyms, antonyms, hypernyms, hyponyms<br>- Any relevant note improving the meaning of business terms | - Glossaries/dictionaries/thesaurus<br>- Code lists<br>- Taxonomies<br>- Ontologies | - Meta-data |
| **3** | Information modeling | - Data at rest and in motion (flows/exchange)<br>- Conceptual, data objects, attributes and relationships<br>- Data provenance and lineage<br>- Data products and services<br>- Process and roles<br>- Responsibility matrix<br>- Systems and applications used | - Interlinked conceptual, logical and physical models<br>- System architecture diagrams with data flows at each interface<br>- List of system/application CRUD actions with corresponding data sets<br>- Dataset provenance and lineage reports<br>- Data service descriptions with corresponding system/application interfaces<br>- Process diagrams (with roles) down to individual tasks linked to corresponding data transformations<br>- Links across data, process, application, ... models | - Data Architecture<br>- Data Modelling & Design<br>- Meta-data |
| **4** | Data FAIRification | - FAIR status and gaps for data in scope | - Report displaying FAIR metrics<br>- Gap analysis and governance prescription to improve data FAIRness | - Data Architecture<br>- Data Security<br>- Data Integration & Interoperability<br>- Data Quality<br>- Meta-data |
| **5** | Master data management | - Master data schema design<br>- Master cross-system identification, disambiguation and mapping<br>- Identification of attributes of type “reference data” and linking to corresponding controlled vocabulary<br>- Linking to relevant external standards | - Master data schemas using governed, validated and standardized metadata<br>- Mapping between master/reference data<br>- Mapping between master data schema fields<br>- Mapping with external standards | - Data Architecture<br>- Data Modelling & Design<br>- Reference & Master Data<br>- Meta-data |
| **6** | Reference data management | - Reference list unique identification<br>- Reference list values unique identification at linkage to the many lists they (can) belong<br>- Inter- and intra-list mapping using fuzzy matching (SKOS) | - Reference data structure and content using governed, validated and standardized metadata<br>- Mapping between reference data values<br>- Mapping with external standards | - Data Architecture<br>- Data Modelling & Design<br>- Reference & Master Data<br>- Meta-data |
| **7** | External standard management | - ETL of external standards<br>- Standards adjustment to enterprise needs<br>- Standards structure/schema and content/values mapping | - List of external data or standard providers<br>- Overlap between external sources<br>- Changes made to external standards when onboarded internally<br>- Governance aspects including version, licensing conditions, roles, lifecycle... | - Data Governance<br>- Data Storage & Operations<br>- Data Integration & Interoperability<br>- Meta-data |
| **8** | Data Governance | - Legal/regulatory artifacts and requirements<br>- FAIRification of rules resulting from these policies<br>- Linking rules to atomic data entities<br>- Linking enterprise objectives, roadmap, milestones, resources, constraints, risks to DG principles<br>- Data quality rules<br>- Enterprise policies and guidelines | - Dynamic reports on governance status, resources and activities<br>- Data strategy and how governance is deployed and maintained<br>- List of governance principles, rules, checks and to what digital asset they apply | - Data Governance<br>- Data Security<br>- Documents & Content<br>- Data Quality<br>- Meta-data |

***Table 1: List of different use cases leveraging RDF to capture, integrate knowledge and generate data governance artefacts.***

| InputData Stage | Input Variable | Derivation | Derivation Rule | Output Variable | OutputData Stage |
|---|---|---|---|---|---|
| 4 | RP.AE.AEENDAT<br><br>[End Date of Adverse Event] | NCDS COPY_ELEMENT | AEENDAT = AE.AEENDAT | DR.AE.AEENDAT<br><br>[End Date of Adverse Event] | 5 |
| 4 | RP.AE.AEENTIM<br><br>[End Time of Adverse Event] | NCDS COPY_ELEMENT | AEENTIM = AE.AEENTIM | DR.AE.AEENTIM<br><br>[End Time of Adverse Event] | 5 |
| 5 | DR.AE.AEENDAT<br><br>[End Date of Adverse Event] | NCDS DTC | Convert date variable AE.AEENDAT and time variable AE.AEENTIM | DR.AE.AEENDTC<br><br>[End Date/Time of Adverse Event] | 5 |
| 5 | DR.AE.AEENTIM<br><br>[End Time of Adverse Event] | NCDS DTC | Convert date variable AE.AEENDAT and time variable AE.AEENTIM | DR.AE.AEENDTC<br><br>[End Date/Time of Adverse Event] | 5 |
| 5 | DR.AE.AEENDTC<br><br>[End Date/Time of Adverse Event] | NCDS DTN | Convert AE.AEENDTC to numeric ISO 8601 format. | DR.AE.AEENDTN<br><br>[End Date/Time of Adverse Event] | 5 |
| 5 | DR.AE.AEENDTN<br><br>[End Date/Time of Adverse Event] | NCDS STUDY_DAY | Calculate the study day from a given start date. The start day is day 1, then day prior to this is day -1. There is no day 0. If AE.AEENDTN is on or after AE.RFSTDTN then (date portion of AE.AEENDTN) - (date portion of AE.RFSTDTN )+ 1... | DR.AE.AEENDY<br><br>[Study Day of End of Adverse Event] | 5 |
| 5 | DR.AE.AEENDY<br><br>[Study Day of End of Adverse Event] | NCDS COPY_ELEMENT | AEENDY = AE.AEENDY | SD.AE.AEENDY<br><br>[Study Day of End of Adverse Event] | 6 |
| 5 | DR.AE.AEENDY<br><br>[Study Day of End of Adverse Event] | NCDS COPY_ELEMENT | AEENDY = DR.AE.AEENDY | AD.ADAE.AEENDY<br><br>[Study Day of End of Adverse Event] | 7 |

***Table 2: Entities retrieved using the SPARQL query (**Figure 10**).***

# *Supplementary Information*

## 12. Framework Technology: source data capture, transformation, and validation

### 12.1 Capturing Knowledge using Excel Enrichers

The process of capturing and serialising the information from business discussions using Excel Enricher templates can be achieved by filling input several dedicated tabs (specification or Spec tabs and quality checks or QC tabs) as follows

1) Definition knowledge origin: Description of captured knowledge and the author who translates the knowledge. Reference to the origin of the knowledge (organisation, name of the standard, website) (Spec tabs "step 1 - config").
    a) Unique concept definition: Capture of a discrete list of individual concepts using an identifier (URI), preferred label (the most frequent label used by business), optionally alternate labels and a description (Spec tabs "step 2 - unique concepts").
    b) Concept classification and contextualisation: Tag each concept with a UO Class (OWL Class) representing a key dimension within the EA framework. For example, the concept "Setup a clinical study" will be tagged with the Class "process" (Spec tabs "step 2 - unique concepts").
2) Concept linkage: Link concepts (URIs) to each other using standardised relationships/predicates (also URIs) that exist within the governed predicates list within the UO (Spec tabs "additional properties").
3) Quality steps: Verify transformations (QC tab)
    a) valid and unique URIs
    b) locally unique preferred labels (within the file)
    c) proper definitions of linked concepts
4) Transform information: execute the sequence of Power Queries and perform a quality analysis of the transformation (QC tab). This step checks the data structure (if it is SKOS, OWL classes) and syntax.

Knowledge captured within the Excel Enricher is serialised into TTL format using the inputted specifications (Specs tabs). Data can be loaded into the EKG three ways: either using the TURTLE Loader application, the "drag and drop" functionality in the DGSS portal, or through the ETL pipeline in case further integration or validation is needed.

### 12.2 Semantic ETL pipeline for systems and datasets

#### 12.2.1 Step 1: Data Extract & Serialisation (E)

We start by normalising the serialisation format. Often, after capturing the required information to model a use-case, we end up collecting structured information in a varying set of formats. This information can be files of multiple formats, such as CSV (for example, the excel enrichers), UML (unified modelling standards), JSON and XML, or it can exist in different relational or non-relational databases, such as SQL and MongoDB. As a result of the serialisation step, the data will be available in a Pandas DataFrame (for Python packages).

### 12.2.2 Step 2: Semantic Validation & Transformation (T)

In this step, our algorithmic methods begin by performing a set of standard value transformations, followed by semantic-related validations and transformations, leading to the generation of a semantic model entirely integrated and compliant with the overarching logical model. In this regard, while modelling a dataset, it is essential to build a SHACL schema for each dataset, which will describe the resultant model semantic model within the enterprise resources.

All details associated with the semantic transformations (such as code, language, execution environment) are recorded in the knowledge graph and associated with the original input dataset and the output semantic model.

1. Dataset restructuring and standard value transformations: Value transformations can be seen as functions that take a dataset row as their input (from the normalised tabular dataset) and return it with transformed values. Commonly, each field/feature represented in a column is used to generate the output after simple transformations; however, transformations can also create new ("computed") columns. Simple examples of transformations include trimming of white space, changing letter cases and parsing numbers and dates. Complex transformations can exploit the full capabilities of programming languages (e.g., JavaScript / NodeJS, Python). These transformations are often used to clean data and adequately expose the data content based on pre-defined table schemas.
2. Semantic transformations, validations and serialisation: In this step, we implement the (meta)data management actions detailed in the knowledge integration section of the framework methodology. A table schema (SHACL schema) is generated for each previously extracted and serialised tabular dataset. This schema semantically describes the dataset and allows us to validate the dataset's content (Classes, Properties and Values).
    a. Structural validation of a table schema: The SHACL schema reflects the dataset table schema and augments it by appending the semantic contexts (URI prefixes) for its content (Classes, Properties and Values) and defining the SHACL attributes for subsequent semantic validation. The structural validation of the table schema ensures that it correctly renders the dataset information into valid RDF triples:
        i. It defines a primary key column that contains the instantiation of a Class (for example, an 'ID' column). The remaining columns must relate to the primary key column.
        ii. It uses the non-primary key column headers to define the Property by which the primary key's instances relate to the corresponding column values.
        iii. It defines the data type of the column values by either instantiating a Class (similarly to the primary key column) or by describing it as a literal/string.
        iv. It verifies the correctness of the URIs' syntax.
    b. Semantic validation of classes and properties: The SHACL schema is then used to validate the dataset Classes and Properties against those defined (using RDFS and OWL) within the enterprise RDF model. This step ensures proper integration of the dataset with the enterprise knowledge base. The following types of semantic validations are performed:
        i. Validation of cardinality: Each field may be constrained to occur at specified cardinality. The minCount and maxCount attributes enable us to declare required vs optional fields and multi-valued fields with specified minimal and maximal number of entries. Multi-valued fields without an upper bound on the maximum number of entries can be specified, leaving the maxCount attribute empty.
        ii. Validation of literals: Literal values can be validated based on the type of the column as specified in the dataset definition, e.g., number, date-time. Both JSON schema and SHACL attributes can ensure the value conforms to the specification.
        iii. Validation of referential integrity: Values that instantiate classes may be already represented in the knowledge graph, which is the case of reference values. Given access to the knowledge

graph, a SHACL validator can verify that the referenced URI exists within the target graph and that it is associated with a reference list conforming to the appropriate model (schema). This form of referential integrity checking is crucial to ensuring that the knowledge graph is maximally connected (free of "dangling pointers"). Therefore SHACL-based integrity validations ensure the generation of adequately formed logical models. An interesting consequence of the referential integrity checking implementation at the knowledge graph level is that it can detect a "new class" within a dataset. This validation alerts the necessity to fully characterise this "new class" within the glossary, the upper ontology, and describe it with a semantic model. This alert occurs since these resources fail to integrate this "new class" according to their respective models. It is worth noting that the SHACL schema does not natively support validating that a value is a member of a class with a large number of members. The "enum" construct of the SHACL schema does not scale well, considering that some of our use cases require validation against classes with millions of members.

iv. Validation of data at record level: Despite the clear relevance and potential benefits of such implementation, we do not yet validate all data at the record level. This record-level SHACL validation is not technically challenging, and it can be easily implemented within our pipeline. However, the curation effort required to deal with validation failures will be significant.

c. Generation of a conformant RDF model: For a given dataset, once a valid SHACL-schema is generated and the semantic validations are completed, the dataset is transformed into RDF triples that conform to the enterprise RDF model.

### 12.2.3 Step 3: Load logical model (L)

At this stage, the data is available as a set of conformant RDF triples and can be directly loaded to the triple store, using the TURTLE Loader, the "drag and drop" functionality in the DGSS portal (see Appendix 4), or through database-specific Python methods used on this pipeline. The previous sequence of transformation and validation steps guarantees complete integration with the existing knowledge base.

## 12.3 Framework Technology: supporting applications

### 12.3.1 DGSS Portal

The DGSS Portal is created using React, a flexible component based Javascript library for web applications. The portal runs on Apache Tomcat, the most used open-source web application server

The intention from the start was to create a data-driven portal where the entire structure and content of the portal could be dynamically generated using data stored in the graph database. This approach allows an agile and collaborative approach to content definition. The only pages that are not dynamically generated are used for searching the content of the graph database, to allow users to upload and test RDF generated by enrichers, and an interactive wizard assisting in the creation of URIs based on business data.

Page content is stored in triples using Markdown format. Markdown was selected as it allows richly formatted content to be defined in a terse and relatively human-readable format. The markdown-to-jsx component was chosen for rendering content, as this allowed Markdown to be extended with additional tags, so the content is also able to define more complex layouts, and the dynamically generated tables and knowledge graphs hosted on the pages. These tables and graphs are described in more detail in the DGSS Publisher section.

Additionally, the SPARQL queries used for populating the data-driven tables and knowledge graphs are stored as triples. This is described in the DGSS R1 section.

### 12.3.2 DGSS Publisher

The DGSS Publisher is a standalone JavaScript-based widget providing several rendering modes of triples. Once configured, it can autonomously get triples from SPARQL queries (dynamic) or sparql-json files (static) and display the result as table (based on https://datatables.net/), network (based on https://visjs.org/) or ID card (inspired from the SPARQL keyword DESCRIBE).

Rendering modes are the following:

- Table: display triples in a tabular format
- Mapper: a specific tabular format with feature to help mapping concepts between other concepts
- Feedback: a specific tabular format with a comment feature stored in triples
- Compare: a specific tabular format aiming to compare concepts between concepts
- Network: display triples in a network format
- Network Explorer: display one concept in a network format with a navigation feature to linked concepts
- Describe: display one concept identity as an ID card with all its properties

### 12.3.3 DGSS R1

The DGSS R1 is a simple Java web application aiming to redirect URL based on rules and is used as an API for the DGSS Portal. The redirection feature can be used to forward a company-based internal URL to and external public URL (like a source of truth). The API feature encapsulate and hide SPARQL queries that are used by the DGSS Portal to display information

The configuration of rules and the SPARQL queries used by the API is stored as triples in the triple store.

### 12.3.4 DGSS URI builder

In order to enforce governance around URI syntax, we provide a “URI builder” application for everyone to build URI step-by-step.

### 12.3.5 DGSS Table builder

This tool helps presenting data as a table based on the entity properties. This tool automatically builds SPARQL query that extracts the required properties from the RDF store.

## 12.4 TurtleLoader

The TurtleLoader is a GUI, Java-based, desktop client, aiming to facilitate and accelerate the ingestion of RDF and Turtle files by the triple store. The first of its feature is to enforce a unified naming convention for graph name (file name-based, used in N-Quad). By enforcing a naming convention, it makes the update of existing triples easier (existing graph name are cleared before adding new triples). Duplicated triples are also removed automatically, spog-based. It can also generate predicates based on the existence of an owl:inverseOf relationship.

The three primary features are:

- Import: RDF or Turtle files can be sent to the triple store.

- Enrich: SPARQL CONSTRUCT/INSERT queries can be sent to the triple store, output is new triples.
- Generate: SPARQL SELECT queries can be sent to the triple store, output is sparql-json useable by a web browser.

## 12.5 RDF Generate (Java)

An alternative to the Excel based enrichers was needed for certain operational data sources, where the number of records can run in to 100's of thousands, or where the source could be a REST API, a relational database or many thousands of Excel files.

For these cases, Java was used to create an executable program to allow large datasets to be converted to RDF.

This program has three distinct stages:

- Stage 1: Access the data sources and extracts the relevant information. For example, this may call a REST based web-service, or access an Oracle database using JDBC. The retrieved data can be stored in one of several formats, for example JSON or CSV.
- Stage 2: Each of the data sources is converted into Turtle format using a SPARQL template. For this step the SPARQL-Generate API was chosen. This Java API performs well over larger datasets, supports a variety of file formats, and offers a variety of iterators and functions that help with the process.
- Stage 3: Where applicable, the generated Turtle from several data sources can be combined into a single file, for ease of loading and maintenance.

To add a new data-source, the program requires configuration to be added in JSON format. This identifies the data source, its type (for example, a database table), data source specific information (for example, a connection string, credentials, table name and required fields) and the name of the template used to do the conversion.

Additionally, a SPARQL-Generate template is written for specifically for this data source. This converts the data downloaded from the data source into Turtle format (https://ci.mines-stetienne.fr/sparql-generate/).

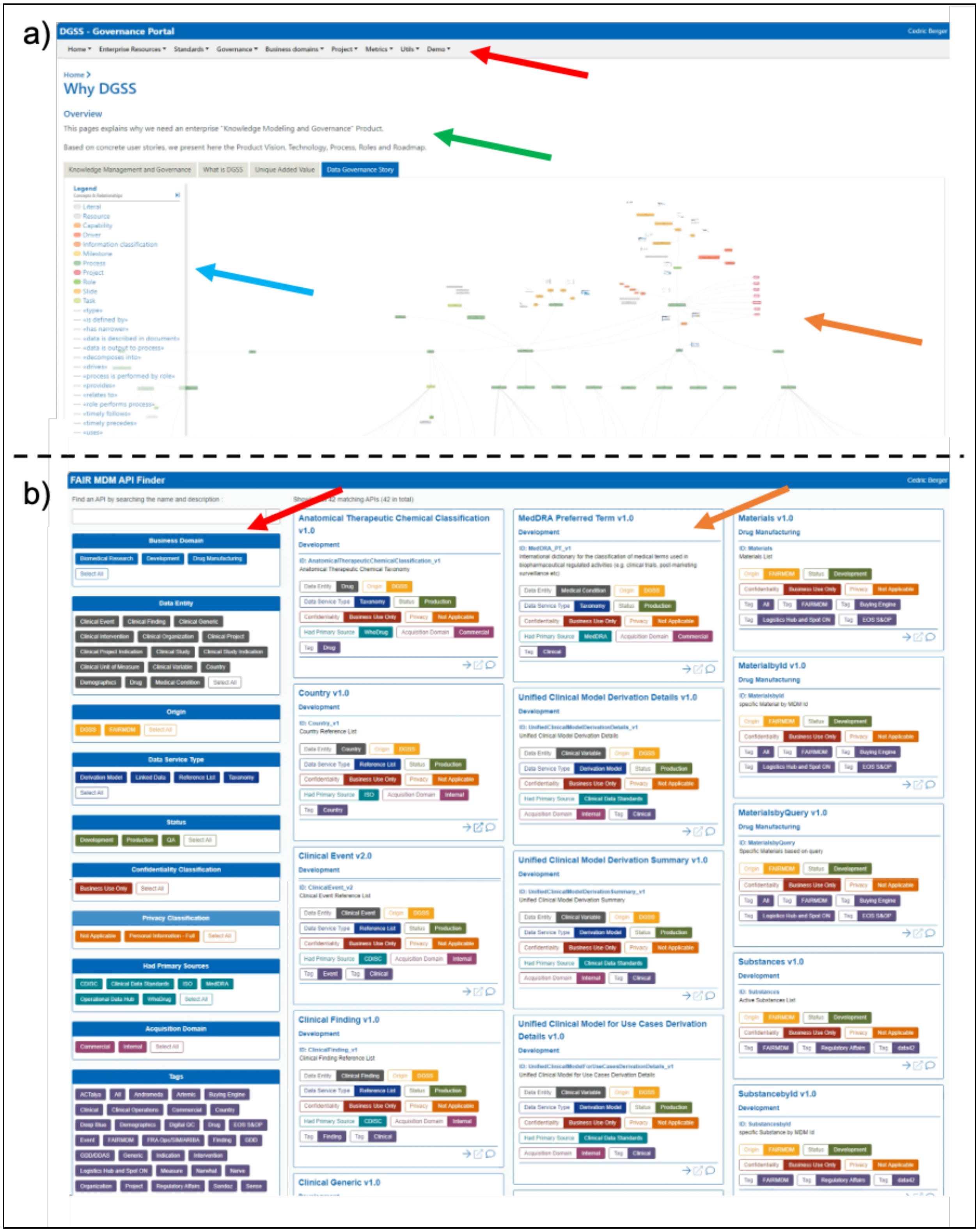

***Figure S1: Screenshot of the DGSS portal (a) and the FAIR MDM API portal (b).*** *A top menu (red arrow) shows high-level categories that group similar DGSS pages. Pages in DGSS consist in HTML content (green arrow) and tabular or graphical (orange arrow) representation of KG content. Graph visualisation are always showing a legend explaining upper ontology entities displayed on the screen (blue arrow). The HTML part provides a narrative explanation about the use case, how the KG is adding value and what it represents.*